\documentclass[11pt]{article}
\usepackage{graphicx}
\usepackage{amssymb}
\usepackage{amscd}
\usepackage{mathrsfs}
\usepackage{longtable,lscape}
\usepackage{amsthm}
\usepackage{amsfonts}
\usepackage{amsmath}
\usepackage{bbm}
\usepackage{float}
\usepackage{url}

\newcommand{\captionfonts}{\footnotesize}
\makeatletter  
\long\def\@makecaption#1#2{%
  \vskip\abovecaptionskip
  \sbox\@tempboxa{{\captionfonts #1: #2}}%
  \ifdim \wd\@tempboxa >\hsize
    {\captionfonts #1: #2\par}
  \else
    \hbox to\hsize{\hfil\box\@tempboxa\hfil}%
  \fi
  \vskip\belowcaptionskip}
\makeatother 
\begin{document}
\title{Quantum Structure in Cognition \\ Origins, Developments, Successes and Expectations}
\author{Diederik Aerts$^1$ and Sandro Sozzo$^{2}$
		 \vspace{0.5 cm} \\ 
        \normalsize\itshape
        $^1$ Center Leo Apostel for Interdisciplinary Studies \\
        \normalsize\itshape
        and, Department of Mathematics, Brussels Free University \\ 
        \normalsize\itshape
         Krijgskundestraat 33, 1160 Brussels, Belgium \\
        \normalsize
        E-Mail: \url{diraerts@vub.ac.be}
          \vspace{0.5 cm} \\ 
        \normalsize\itshape
        $^2$ School of Management and IQSCS, University of Leicester \\ 
        \normalsize\itshape
         University Road, LE1 7RH Leicester, United Kingdom \\
        \normalsize
        E-Mail: \url{ss831@le.ac.uk, ssozzo@vub.ac.be}
           \\
              }
\date{}
\maketitle
\begin{abstract}
\noindent
We provide an overview of the surprising results we have attained in the last decade on the identification of quantum structures in cognition and, more specifically, in the formalization and representation of natural concepts. 
We firstly discuss the quantum foundational reasons that led us to investigate the mechanisms of formation and combination of concepts in human mind, starting from the empirically observed deviations from classical logical and probabilistic structures.
We then develop our quantum-theoretic perspective in Fock space which allows successful modeling of various sets of cognitive experiments collected by different scientists, including ourselves. In addition, we formulate a unified explanatory hypothesis for the presence of quantum structures in cognitive processes, and discuss our recent discovery of further quantum aspects in concept combinations, namely, `entanglement' and `indistinguishability'. We finally illustrate perspectives for future research.
\end{abstract}

\section{The combination problem in concept theory\label{intro}}
That concepts exhibit aspects of `contextuality', `vagueness' and `graded typicality' was already known in the seventies since the 
 investigations of Eleanor Rosch \cite{r1973}. These studies questioned  explicitly the traditional view that `concepts are containers of instantiations' and, additionally, although not explicitly stated, there was already the suspicion that `the human mind combines concepts not following the algebraic rules of classical logic even if the combinations are conjunctions or disjunctions'. In particular, conceptual gradeness led scholars to introduce elements of probability theory in structuring and representing concepts. A possible way to at least preserve a set-theoretical basis was the fuzzy set approach \cite{z1989}. According to this proposal, concepts would combine in such a way that the conjunction of two concepts satisfies the `minimum rule of fuzzy set conjunction' and the disjunction of two concepts satisfies the `maximum rule of fuzzy set disjunction'. In this way, one could still maintain that `concepts can be represented as (fuzzy) sets and combine according to set-theoretic rules'. However, a whole set of experimental findings in the last thirty years revealed that the latter does not hold, thus raising the so-called `combination problem'.

(i) `Guppy effect'. Osherson and Smith measured the `typicality' of specific exemplars with respect to the concepts {\it Pet} and {\it Fish} and their conjunction {\it Pet-Fish} \cite{os1981}, and they found that people rate an exemplar such as {\it Guppy} as a very typical example of {\it Pet-Fish}, without rating {\it Guppy} as a very typical example neither of {\it Pet} nor of {\it Fish} (`Pet-Fish problem').\footnote{In a typicality measurement, subjects are asked to choose the exemplar they consider as the most typical example of a given concept, hence they have to pick the best example in a list of items.} Interestingly enough, this guppy effect violates the minimum rule of fuzzy set conjunction.

(ii) `Overextension and underextension effects'. Hampton measured the `membership weight' of several exemplars with respect to specific pairs of concepts and their conjunction \cite{h1988a} and disjunction \cite{h1988b}, finding systematic deviations from fuzzy set modeling. Adopting his terminology, if the membership weight of an exemplar $x$ with respect to the conjunction `$A \ {\rm and} \ B$' of two concepts $A$ and $B$ is higher than the membership weight of $x$ with respect to one concept (both concepts), we say that the membership weight of $x$ is `overextended' (`double overextended') with respect to the conjunction (briefly, $x$ is overextended with respect to the conjunction). If the membership weight of an exemplar $x$ with respect to the disjunction `$A \ {\rm or} \ B$' of two concepts $A$ and $B$ is less than the membership weight of $x$ with respect to one concept (both concepts), we say that the membership weight of $x$ is `underextended' (`double underextended') with respect to the disjunction (briefly, $x$ is underextended with respect to the disjunction). We have recently performed a similar cognitive test on conceptual conjunctions of the form `$A$ and $B$' \cite{s2014b,asv2014}, detecting systematic overextension and also double overextension.\footnote{In a membership measurement, subjects are asked to decide whether a given exemplar $x$ is a member of a given concept $A$. When many subjects are involved in the measurement, a membership weight $\mu(A)$ can be defined for $x$ as a large number limit of the relative frequency of positive answers.} 

(iii) `Deviations from classicality in conceptual negation'. More recently, Hampton measured the membership weights of many exemplars with respect to specific pairs of concepts and their conjunction, e.g., {\it Tools Which Are Also Weapons}, and also conjunction when the second concept is negated, e.g., {\it Tools Which Are Not Weapons} \cite{h1997}. He detected overextension in both types of conjunctions, as well as deviations from classical logical behaviour in conceptual negation. We have recently performed a more general cognitive test \cite{s2014b,asv2014}, detecting systematic overextension, double overextension and violation of classical logic negation rules in conceptual conjunctions of the form `$A$ and not $B$', `not $A$ and $B$' and `not $A$ and not $B$'.

(iv) `Borderline contradictions'. Alxatib and Pelletier asked human subjects to estimate the truth value of a sentence such as ``$x$ is tall and not tall'' for a given person $x$ who was showed to the eyes of the subjects \cite{ap2011}. They found that a significant number of subjects estimated such a sentence as true, in particular, for borderline cases.\footnote{A borderline contradiction can be formalized as a sentence of the form $P(x) \land \lnot P(x)$, for a vague predicate $P$ and a borderline case $x$, e.g., the sentence ``Mark is rich and Mark is not rich''.} 

Difficulties (i)--(iv) entail, in particular, that the formation and combination rules of human concepts do not generally follow the laws of classical (fuzzy set) logic \cite{os1981,h1988a,h1988b,h1997}. Moreover, the corresponding experimental data cannot generally be modeled in a single classical probability space satisfying the axioms of Kolmogorov, which we proved in various articles \cite{s2014b,a2009a,ags2013,s2014a}.

Our investigation of the above `deviations from classicality'\footnote{By the locution `deviation from classicality' we actually mean that 
 classical logical and probabilistic structures, i.e. the most traditional models of cognition, cannot account for the experimentally observed patterns} in conceptual combinations can be traced back to our studies on the axiomatic and operational foundations of quantum physics and the origins of quantum probability (see, e.g., \cite{a1986}). We recognized that any decision process, e.g., a typicality measurement, or a membership estimation, involves a `transition from potential to actual', in which an outcome is actualized from a set of possible outcomes as a consequence of a contextual interaction (of a cognitive nature) of the subject with the conceptual situation that is the object of the decision. Hence, human decision processes exhibit deep analogies with what occurs in a quantum measurement process, where the measurement context (of a physical nature) influences the measured quantum particle in a non-deterministic way. Quantum probability -- which is able to formalize this `contextually driven actualization of potential', not classical probability -- which only formalizes lack of knowledge about actuality -- can conceptually and mathematically cope with this situation underlying both quantum and conceptual realms \cite{aa1995}. 

The second step of our research was the elaboration of a `State Context Property' (SCoP) formalism to abstractly represent any type of entity, e.g., a conceptual entity, in terms of its states, contexts and properties. In SCoP, a concept is represented as an `entity being in a specific state and changing under the influence of a cognitive context', rather than as a `container of instantiations', and we were able to provide a quantum-theoretic model in Hilbert space that successfully describes the guppy effect \cite{ag2005a,ag2005b} (Section \ref{contextuality}). 

The successive development of our research was the employment of the mathematical formalism of quantum theory in Fock space to model the overextension and underextension of membership weights measured in \cite{h1988a,h1988b}. These effects can be described in terms of genuine quantum aspects, like `interference', `superposition' and `emergence' \cite{a2009a,ags2013,a2009b,abgs2013}. This quantum-mechanical model was successfully applied to describe borderline contradictions \cite{s2014a}. More recently, we extended the model to incorporate conceptual negation, thus faithfully representing the above mentioned experiments by ourselves on concept conjunctions and negations \cite{s2014b,asv2014} (Section \ref{fockspace}). 

Our results allowed us to put forward a unifying explanatory hypothesis for this whole set of experimental findings in human cognition, namely, that human thought is guided by two simultaneous processes -- `quantum conceptual thought', whose nature  is `emergence', and `quantum logical thought', whose nature is `logic' \cite{asvIQSA2}.  Our investigations indicate that the former generally prevails over the latter, and that the effects, paradoxes, contradictions, fallacies, experimentally detected in human cognition can be considered as expressions of this dominance, rather than  `biases' of the human mind. More recently, we received a further crucial confirmation of this two-layered structure in human thought, namely the stability of the deviation from classical probabilistic rules we detected in \cite{asv2014} (Section \ref{explanation}).

Our quantum-theoretic perspective also accounts for two recent experimental results we obtained, namely, the identification of `quantum entanglement' in the conceptual combination {\it The Animal Acts} \cite{as2011,as2014} (Section \ref{entanglement}) and the detection of `quantum indistinguishability of Bose-Einstein type' in specific combinations of identical concepts, such as {\it Eleven Animals} \cite{asvIQSA1} (Section \ref{identity}). These discoveries are also important, in our opinion, from the point of view of the foundations of quantum physics, since they can shed new light on two mysterious aspects of the microscopic world -- entanglement and indistinguishability. 

In this review article, we present the above results by basically following a historical reconstruction, though justified and restructured in a unitary and more general rational framework. We conclude our paper with some epistemological remarks on the role and interpretation of our quantum-theoretic perspective within the domain of cognitive modeling, and with some hints for future developments (Section \ref{conclusions}).

\section{The first steps: potentiality and contextuality in decision processes\label{contextuality}}
The first move towards the development of a quantum-theoretic perspective in cognition came from our former research on the mathematical and conceptual foundations of quantum physics. In particular, we were guided by our studies on:

(i)  the identification of quantum structures outside the microscopic world, e.g., in the cognitive situation of the liar paradox \cite{abs1999,aabg2000};

(ii) the recognition of the existence of deep analogies between quantum particles and conceptual entities with respect to `potentiality' and `contextuality';

(iii)  the role played by quantum probability in formalizing experimental situations where these aspects of potentiality and contextuality occur.

It is well known from quantum physics that, in a quantum measurement process, the measurement context influences the quantum entity that is measured in a non-deterministic way, actualizing one outcome in a set of possible measurement outcomes, as a consequence of the interaction between the quantum entity and the measurement context. Suppose now that a statistics of measurement outcomes is collected after a sequence of many repeated measurement processes on  an arbitrary entity, and such that (i) the measurement actualises properties of the entity that were not actual before the measurement started, (ii) different outcomes and actualisations are obtained probabilistically. What type of probability can formalize such experimental situation? It cannot be classical probability,  because classical probability formalizes lack of knowledge about actual properties of the entity that were already actual before the measurement started. We proved many years ago that a situation where context actualizes potential properties can instead be represented in a suitable quantum probabilistic framework \cite{a1986}.

What about a human decision process? Well, we realized that a decision process is generally made in a state of genuine potentiality, which is not of the type of a lack of knowledge of an actuality. The following example may help to illustrate this point. In \cite{aa1995}, we considered a survey including the question ``are you a smoker or not?''. Suppose that 21 participants over a whole sample of 100 participants answered  `yes' to this question. We can then consider 0.21 as the probability of finding a smoker in this sample of participants. However, this probability is obviously of the type of a `lack of knowledge about an actuality', because each participant `is' a smoker or `is not' a smoker before the property has been tested, hence before the experiment to test it -- the survey -- starts. Suppose that we now consider the question ``are you for or against the use of nuclear energy?'', and that 31 participants answer they are in favor. In this case, the resulting probability $0.31$ is `not' of the type of `lack of knowledge about an actuality'. Indeed, it is very plausible for this type of question that some of the participants had no opinion about  it before the survey,  and hence for these participants the outcome was influenced by the context at the time the question was asked, including the specific conceptual structure of how the question was formulated. This is how context plays an essential role whenever the human mind is concerned with outcomes of experiments of a cognitive nature. We showed that the first type of probability, i.e. the type that models a `lack of knowledge about an actuality', is classical, and that the second type is non-classical and, possibly, quantum \cite{a1986}.

The effect due to role that context plays on a conceptual entity is equally fundamental than  the effect due to the actualizing of potentialities during a decision process. Exactly as in a quantum measurement the measurement context changes the state of the quantum entity that is measured, in a decision process the cognitive context changes the state of the concept  \cite{ag2005a,ag2005b}. For example, in our modeling of the concept {\it Pet}, we considered the context $e$ expressed by {\it Did you see the type of pet he has? This explains that he is a weird person}, and found that when participants in an experiment were asked to rate different exemplars of {\it Pet}, the scores for {\it Snake} and {\it Spider} were very high in this context. In our perspective, this is explained by the existence of different states for the concept {\it Pet}, where we use the notion of `state' in the same way as it is used in quantum theory, but also as it is used in ordinary language, i.e. `the state of affairs', meaning `how the affairs will react on different measurement contexts. We call `the state of {\it Pet} when no specific context is present', its ground state $\hat p$. The context $e$ then changes the ground state $\hat p$ into a new state $p_{weird\ person\ pet}$. Typicality, in our perspective, is an observable semantic quantity, which means that it takes different values in different states. Hence, in our perspective the typicality variations as encountered in the guppy effect are due to changes of state of the concept {\it Pet} under influence of a context. More specifically, the conjunction {\it Pet-Fish} is {\it Pet} under the context {\it Fish}, in which case the ground state $p$ of {\it Pet} is changed into a new state $p_{Fish}$. The typicality of {\it Guppy}, being an observable semantic quantity, will be different depending on the state, and this explains the high typicality of {\it Guppy} in the state $p_{Fish}$ of {\it Pet}, and its normal typicality in the ground state $p$ of {\it Pet} \cite{ag2005a}.

We developed this approach in a formal way, and called the underlying mathematical structure a `State Context Property' (SCoP) system \cite{ag2005a}. To build SCoP for an arbitrary concept $S$ we introduce three sets, namely, the set $\Sigma$ of states, denoting states by $p, q, \ldots$, the set ${\mathcal M}$ of contexts, denoting contexts by $e, f, \ldots$, and the set ${\mathcal L}$ of properties, denoting properties by $a, b, \ldots$. The `ground state' $\hat{p}$ of the concept $S$ is the state where $S$ is not under the influence of any particular context. Whenever $S$ is under the influence of a specific context $e$, a change of the state of $S$ occurs. In case $S$ was in its ground state $\hat{p}$, the ground state changes to a state $p$. The difference between states $\hat{p}$ and $p$ is manifested, for example, by the typicality values of different exemplars of the concept, and the applicability values of different properties being different in the two states $\hat{p}$ and $p$. Hence, to complete the mathematical construction of SCoP, also two functions $\mu$ and $\nu$ are introduced. The function $\mu: \Sigma \times {\mathcal M} \times \Sigma \longrightarrow [0, 1]$ is defined such that $\mu(q,e,p)$ is the probability that state $p$ of concept $S$ under the influence of context $e$ changes to state $q$ of concept $S$. The function $\nu: \Sigma \times {\mathcal L} \longrightarrow [0, 1]$ is defined such that $\nu(p,a)$ is the weight, or normalization of applicability, of property $a$ in state $p$ of concept $S$. With these mathematical structures and tools the SCoP formalism copes with both `contextual typicality' and `contextual applicability'.

We likewise built an explicit quantum-mechanical representation in a complex Hilbert space of the data of the experiment on {\it Pet} and {\it Fish} and different states of {\it Pet} and {\it Fish} in different contexts explored in \cite{ag2005a}, as well as of the concept {\it Pet-Fish} \cite{ag2005b}. In this way, we were able to cope with the pet-fish problem illustrated in Section \ref{intro}, (i).

The analysis above already contained the seeds of our quantum modeling perspective for concept combinations -- in particular, the notion of state of a concept marked the departure from the traditional idea of a concept as a set, eventually fuzzy, that contains instantiations. However, this analysis was still preliminary, and a general quantum-mechanical modeling required further experimental and theoretic steps, as it will be clear from the following section.

\section{Modeling concept combinations in Fock space\label{fockspace}} 
We present here our quantum modeling perspective in Fock space for the combination of two concepts. It is successful in describing the classically problematical results illustrated in Section \ref{intro}, (ii) (concept conjunction and disjunction), (iii) (concept negation) and (iv) (borderline contradictions).

Let us firstly consider the membership weights of exemplars of concepts and their conjunctions/disjunctions measured by Hampton   \cite{h1988a,h1988b}. He identified systematic deviations from classical (fuzzy) set  conjunctions/disjunctions, an effect known as `overextension' or `underextension' (see Section \ref{intro}). We showed in \cite{a2009a} that a large part of Hampton's data cannot be modeled in a classical probability space satisfying the axioms of Kolmogorov \cite{k1933,p1989}. For example, the exemplar {\it Mint} scored in \cite{h1988a} the membership weight 
$\mu(A)=0.87$ with respect to the concept {\it Food},
$\mu(B)=0.81$ with respect to the concept  {\it Plant},
and $\mu(A \ {\rm and } \ B)=0.9$ with respect to their conjunction {\it Food And Plant}. Thus, the exemplar \emph{Mint} exhibits overextension with respect to the conjunction \emph{Food And Plant} of the concepts \emph{Food} and \emph{Plant}, and no classical probability representation exists for these data. More generally, the membership weights $\mu(A), \mu(B)$ and $\mu(A\ {\rm and}\ B)$ of the exemplar $x$ with respect to concepts $A$, $B$ and their conjunction `$A$ and $B$', respectively, can be represented in a classical Kolmogorovian probability model if and only if they satisfy the following inequalities \cite{s2014b,a2009a}
\begin{eqnarray} \label{ineq01}
&\mu(A\ {\rm and}\ B)-\min(\mu(A),\mu(B)) \le 0 \\ 
& \mu(A) + \mu(B) - \mu(A\ {\rm and}\ B) \le 1\label{ineq02}
\end{eqnarray}
A violation of (\ref{ineq01}) entails, in particular, that the minimum rule of fuzzy set conjunction does not hold, as in the case of {\it Mint}. A similar situation occurs in the case of disjunctions. We showed in \cite{a2009a} that a large part of Hampton's data cannot be modeled in a classical Kolmogorovian probability space. For example,  
the exemplar  {\it Sunglasses} scored in \cite{h1988b}
the membership weight $\mu(A)=0.4$ with respect to the concept {\it Sportswear}, $\mu(B)=0.2$ with respect to the concept
{\it Sports Equipment}, and
$\mu(A \ {\rm or} \ B)=0.1$
with respect to their disjunction
{\it Sportswear Or Sports Equipment}. Thus, the exemplar 
\emph{Sunglasses} exhibits underextension with respect to the disjunction \emph{Sportswear Or Sports Equipment} of the concepts \emph{Sportswear} and \emph{Sports Equipment}, and no classical probability representation exists for these data. More generally, the membership weights $\mu(A), \mu(B)$ and $\mu(A\ {\rm or}\ B)$ of the exemplar $x$ with respect to concepts $A$, $B$ and their disjunction `$A$ or $B$', respectively, can be represented in a classical Kolmogorovian probability model if and only if they satisfy the following inequalities \cite{a2009a}
\begin{eqnarray} \label{maxdeviation}
\max(\mu(A),\mu(B))-\mu(A\ {\rm or}\ B)\le 0 \\ \label{kolmogorovianfactordisjunction}
0 \le \mu(A)+\mu(B)-\mu(A\ {\rm or}\ B)
\end{eqnarray} 
A violation of (\ref{maxdeviation}) entails, in particular, that the maximum rule of fuzzy set disjunction does not hold, as in the case of {\it Sunglasses}. 

In a first attempt to elaborate a quantum mathematics model for the data in \cite{h1988a} and \cite{h1988b} we were inspired by the quantum two-slit experiment.\footnote{In the present paper we use for our modeling purposes the standard quantum-mechanical formalism that is presented in modern manuals of quantum physics (see, e.g., \cite{d1958}). A basic summary of this formalism is contained in the volume including this article \cite{thisvolume}.} Consider, for example, the disjunction of two concepts.  This led us to suggest the following Hilbert space model. One could represent the concepts $A$ and $B$ by the unit vectors $|A\rangle$ and $|B\rangle$, respectively, of a Hilbert space $\cal H$, and describe the decision measurement of a subject estimating whether the exemplar $x$ is a member of $A$ by means of a dichotomic observable represented by the orthogonal projection operator $M$. The probabilities $\mu(A)$ and $\mu(B)$ that $x$ is chosen as a member of $A$ and $B$, i.e. its membership weights, are given by the scalar products $\mu(A)=\langle A| M|A\rangle$ and $\mu(B)=\langle B| M|B\rangle$, respectively. The concept `$A \ \textrm{or} \ B$' is instead represented by the normalized superposition ${1 \over \sqrt{2}}(|A\rangle+|B\rangle)$ in $\cal H$. If $|A\rangle$ and $|B\rangle$ are chosen to be orthogonal, that is, $\langle A| B\rangle=0$, the membership weights $\mu(A), \mu(B)$ and $\mu(A\ {\rm or} \ B)$ of an exemplar $x$ for the concepts $A$, $B$ and `$A \ {\rm or} \ B$' are given by
\begin{eqnarray}
\mu(A)&=&\langle A | M|A\rangle \\
\mu(B)&=&\langle B | M|B\rangle \\
\mu(A \ {\rm or} \ B)&=&{1 \over 2}(\mu(A)+\mu(B))+\Re\langle A|M|B\rangle
\end{eqnarray}
respectively, where $\Re\langle A|M|B\rangle$ is the real part of the complex number $\langle A|M|B\rangle$. The term $\Re\langle A|M|B\rangle$ is called `interference term' in the quantum jargon, since it produces a deviation from the average ${1 \over 2}(\mu(A)+\mu(B))$ which would have been observed in the quantum two-slit experiment in absence of interference. In this way, the deviation from classicality in \cite{h1988a,h1988b} would be due to quantum interference, superposition and emergence, exactly as quantum interference, superposition and emergence are responsible of the deviation from the classically expected pattern in the two-slit experiment.

This `emergence-based' model in Hilbert space succeeded in describing many non-classical situations in \cite{h1988a} and \cite{h1988b}. However, it did not work for various cases, and these were exactly the cases where logic seemed to play a role in the mechanism of conceptual combination. This led us to work out a more general model in Fock space. We present the model in the following. We omit proofs and technical details in the following, for the sake of brevity, inviting the interested reader to refer to the bibliography quoted in this section.

In the case of two combining entities, a Fock space $\mathcal F$ consists of two sectors: `sector 1' is a Hilbert space $\cal H$, while `sector 2' is a tensor product $\cal H \otimes \cal H$ of two isomorphic versions of $\cal H$.  It can be proved that a quantum probability model in Fock space exists for Hampton's data on conjunction and disjunction \cite{a2009a,ags2013}. 

Let us start with the conjunction of two concepts. Let $x$ be an exemplar and let $\mu(A)$, $\mu(B)$ and $\mu(A \ {\rm and} \ B)$  be the membership weights of $x$ with respect to the concepts $A$, $B$ and  `$A \ \textrm{and} \ B$', respectively. Let ${\cal F}={\cal H} \oplus ({\cal H} \otimes {\cal H})$ be the Fock space where we represent the conceptual entities. The states of the concepts $A$, $B$ and $`A \ \textrm{and} \ B'$ are represented by the unit vectors $|A\rangle, |B\rangle \in {\cal H}$ and $|A \ \textrm{and} \ B\rangle \in {\cal F}$, respectively, where
\begin{eqnarray}
|A \  \textrm{and} \ B\rangle=m e^{i\lambda}|A\rangle\otimes|B\rangle+ne^{i\nu}{1\over \sqrt{2}}(|A\rangle+|B\rangle)
\end{eqnarray}
The superposition vector ${1 \over \sqrt{2}}(|A\rangle+|B\rangle)$ describes `$A$ and $B$' as a new emergent concept, while the product vector $|A\rangle\otimes|B\rangle$ describes `$A$ and $B$' in terms of concepts $A$ and $B$. The positive numbers $m$ and $n$ are such that $m^{2}+n^{2}=1$. The decision measurement of a subject who estimates the membership of the exemplar $x$ with respect to the concept `$A \  \textrm{and} \ B$' is represented by the orthogonal projection operator $M\oplus (M \otimes M)$ on ${\cal F}$, where $M$ is an orthogonal projection operator on ${\cal H}$. Hence, the membership weight of $x$ with respect to `$A \  \textrm{and} \ B$' is given by
\begin{eqnarray} \label{AND}
\mu(A \ \textrm{and} \ B)&=&\langle A \ \textrm{and} \ B|M \oplus (M \otimes M)|A \ \textrm{and} \ B \rangle \nonumber \\
&=&m^2\mu(A)\mu(B)+n^2 \left ( {\mu(A)+\mu(B) \over 2}+\Re\langle A|M|B\rangle \right )
\end{eqnarray}
where $\mu(A)=\langle A|M|A\rangle$ and $\mu(B)=\langle B|M|B\rangle$ as above. The term $\Re\langle A|M|B\rangle$ is again the interference term of quantum theory. A solution of (\ref{AND}) exists in the Fock space ${\mathbb C}^{3} \oplus ({\mathbb C}^{3} \otimes {\mathbb C}^{3})$ where this interference term  is given by
\begin{equation}
\Re\langle A|M|B\rangle=
\left\{
\begin{array}{ccc}
\sqrt{1-\mu(A)}\sqrt{1-\mu(B)}\cos\theta & & {\rm if} \  \mu(A)+\mu(B)>1 \\
\sqrt{\mu(A)}\sqrt{\mu(B)}\cos\theta  & & {\rm if} \ \mu(A)+\mu(B)\le 1
\end{array}
\right.
\end{equation}
($\theta$ is the `interference angle'). Coming to the example above, namely, the exemplar {\it Mint} with respect to {\it Food}, {\it Plant} and {\it Food And Plant}, we have that (\ref{AND}) is satisfied with $m^2=0.3$, $n^2=0.7$ and $\theta=50.21^{\circ}$. 

The previous mathematical representation admits the following interpretation. Whenever a subject is asked to estimate whether a given exemplar $x$ belongs to the concepts $A$, $B$, `$A \ {\rm and} \ B$', two mechanisms act simultaneously and in superposition in the subject's thought. A `quantum logical thought', which is a probabilistic version of the classical logical reasoning, where the subject considers two copies of exemplar $x$ and estimates whether the first copy belongs to $A$ and the second copy of $x$ belongs to $B$, and further the probabilistic version of the conjunction is applied to both estimates. But also a `quantum conceptual thought' acts, where the subject estimates whether the exemplar $x$ belongs to the newly emergent concept `$A \ {\rm and} \ B$'.
The place whether these superposed processes can be suitably structured is Fock space. Sector 1 hosts the latter process, while sector 2 hosts the former, while the weights $m^2$ and $n^2$ measure the `degree of participation' of sectors 2 and 1, respectively, in the case of conjunction. 
In the case of {\it Mint}, subjects consider {\it Mint} to be more strongly a member of the concept {\it Food And Plant}, than they consider it to be a member of {\it Food} or of {\it Plant}. This is an effect due to a strong presence of quantum conceptual thought, the newly formed concept {\it Food And Plant} being found to be a better fitting category for {\it Mint} than the original concepts {\it Food} or {\it Plant}. 
And indeed, in the case of {\it Mint}, considering the values of $n^2$ and $m^2$, the combination process mainly occurs in sector 1 of Fock space, which means that emergence prevails over logic.

Let us now come to the disjunction of two concepts. Let $x$ be an exemplar and let $\mu(A)$, $\mu(B)$ and $\mu(A \ {\rm or} \ B)$ be the membership weights of $x$ with respect to the concepts $A$, $B$ and `$A \ \textrm{or} \ B$', respectively. Let ${\cal F}={\cal H} \oplus ({\cal H} \otimes {\cal H})$ be the Fock space where we represent the conceptual entities. The concepts $A$, $B$ and `$A \ \textrm{or} \ B$' are represented by the unit vectors $|A\rangle, |B\rangle \in {\cal H}$ and $|A \ \textrm{or} \ B\rangle \in {\cal F}$, respectively, where
\begin{eqnarray}
|A \  \textrm{or}  \ B \rangle=m e^{i\lambda}|A\rangle\otimes|B\rangle+ne^{i\nu}{1\over \sqrt{2}}(|A\rangle+|B\rangle)
\end{eqnarray}
The superposition vector $\frac{1}{\sqrt{2}}(|A\rangle+|B\rangle)$ describes `$A$ or $B$' as a new emergent concept, while the product vector $|A\rangle\otimes|B\rangle$ describes `$A$ or $B$' in terms of concepts $A$ and $B$. The positive numbers $m$ and $n$ are such that $m^{2}+n^{2}=1$, and they estimate the `degree of participation' of sectors 2 and 1, respectively, in the disjunction case. The decision measurement of a subject who estimates the membership of the exemplar $x$ with respect to the concept `$A \  \textrm{or} \ B$' is represented by the orthogonal projection operator $M \oplus ( M \otimes \mathbbmss{1}+\mathbbmss{1}\otimes M - M \otimes M)$ on ${\cal F}$, where $M$ has been introduced above. We notice that
\begin{equation}
 M \otimes \mathbbmss{1}+\mathbbmss{1}\otimes M - M \otimes M= \mathbbmss{1}-
 (\mathbbmss{1}-M)\otimes(\mathbbmss{1}-M)
\end{equation}
that is,  we have applied de Morgan's laws of logic in sector 2 of Fock space in the transition from conjunction to disjunction. The membership weight of $x$ with respect to `$A \  \textrm{or} \ B$' is given by
\begin{eqnarray} \label{OR}
\mu(A \ \textrm{or} \ B)=\langle A \ \textrm{or} \ B | M \oplus (M \otimes \mathbbmss{1}+\mathbbmss{1}\otimes M - M \otimes M)
|A \ \textrm{or} \ B \rangle \nonumber \\
m^2 \left (\mu(A)+\mu(B)-\mu(A)\mu(B) \right  )+n^2 \left ( {\mu(A)+\mu(B) \over 2}+\Re\langle A|M|B\rangle \right )
\end{eqnarray}
where $\mu(A)=\langle A|M|A\rangle$ and $\mu(B)=\langle B|M|B\rangle$. The term $\Re\langle A|M|B\rangle$ is the interference term. A solution of (\ref{OR}) exists in ${\mathbb C}^{3} \oplus ({\mathbb C}^{3} \otimes {\mathbb C}^{3})$ where this interference term 
is given by
\begin{equation}
\Re\langle A|M|B\rangle=
\left\{
\begin{array}{ccc}
\sqrt{1-\mu(A)}\sqrt{1-\mu(B)}\cos\theta & & {\rm if} \  \mu(A)+\mu(B)>1 \\
\sqrt{\mu(A)}\sqrt{\mu(B)}\cos\theta  & & {\rm if} \ \mu(A)+\mu(B)\le 1
\end{array}
\right.
\end{equation}
Coming to the example above, namely, the exemplar 
{\it Sunglasses} with respect to {\it Sportswear}, {\it Sports Equipment} and {\it Sportswear Or Sports Equipment}, we have that  (\ref{OR}) is satisfied with $m^2=0.03$, $n^2=0.97$ and $\theta=155.00^{\circ}$.

The previous mathematical representation admits the following interpretation. Whenever a subject is asked to estimate whether a given exemplar $x$ belongs to the concepts $A$, $B$, `$A \ {\rm or} \ B$', two mechanisms act simultaneously and in superposition in the subject's thought. A `quantum logical thought', which is a probabilistic version of the classical logical reasoning, where the subject considers two copies of exemplar $x$ and estimates whether the first copy belongs to $A$ or the second copy of $x$ belongs to $B$, and further the probabilistic version of the disjunction is applied to both estimates. And also a `quantum conceptual thought' acts, where the subject estimates whether the exemplar $x$ belongs to the newly emergent concept `$A \ {\rm or} \ B$'. The place whether these superposed processes are structured is again Fock space. Sector 1 hosts the latter process, while sector 2 hosts the former, while the weights $m^2$ and $n^2$ measure the `degree of participation' of sectors 2 and 1, respectively, in the case of disjunction. 
In the case of {\it Sunglasses}, subjects consider {\it Sunglasses} to be less strongly a member of the concept {\it Sportswear Or Sports Equipment}, than they consider it to be a member of {\it Sportswear} or of {\it Sports Equipment}. This is an effect due to a strong presence of quantum conceptual thought, the newly formed concept {\it Sportswear Or Sports Equipment} being found to be a less well fitting category for {\it Sunglasses} than the original concepts {\it Sportswear} or {\it Sports Equipment}. 
And indeed, in the case of {\it Sunglasses}, considering the values of $n^2$ and $m^2$, the combination process mainly occurs in sector 1 of Fock space, which means that emergence aspects prevails over logical aspects in the reasoning process.

Let us then analyze the experiment of Alxatib and Pelletier on borderline contradictions \cite{ap2011}. We proved in \cite{s2014a} that our quantum-theoretic model for the conjunction correctly represents the collected data. Suppose that a large sample of human subjects is asked to estimate the truth values of the sentences ``John is tall'', ``John is not tall'' and ``John is tall and not tall'', for a given subject John showed to the eyes of the subjects. And suppose that the fractions of positive answers are $0.01$,  $0.95$ and $ 0.15$, respectively \cite{ap2011}. This `borderline case' is clearly problematical from a classical logical perspective, and can be modeled in terms of overextension. Indeed, 
let us denote by $\mu(A)$, $\mu(A')$ and $\mu (A \ {\rm and} \ A')$ the probabilities that the sentences ``John is tall'', ``John is not tall'' and ``John is tall and not tall'' are true, and interpret them as membership weights of the exemplar {\it John} with respect to the concepts {\it Tall}, {\it Not Tall} and {\it Tall And Not Tall}, respectively. Then  ({\ref{AND}}) is solved for $m^{2}=0.77$, $n^{2}=0.23$ and $\theta=0^{\circ}$ \cite{s2014a}. The explanation of this behaviour is that the reasoning process of the subject mainly occurs in sector 2 of Fock space, hence 
logical reasoning is dominant, although emergent reasoning is also present, and it is its presence which evoked the name `contradiction' for this situation.

Let us finally come to the experiments on conceptual negation. The first studies on the negation of natural concepts were also performed by Hampton \cite{h1997}. He tested membership weights on conceptual conjunctions  of the form {\it Tools Which Are Not Weapons}, finding overextension and deviations from
Boolean behaviour in the negation. We recently performed a more general cognitive test inquiring into the membership weights of exemplars with respect to conjunctions of the form {\it Fruits And Vegetables}, {\it Fruits And Not Vegetables}, {\it Not Fruits And Vegetables} and {\it Not Fruits And Not Vegetables} \cite{s2014b,asv2014}. Our
   data confirmed significant deviations from classicality
and evidenced a very stable pattern of such deviations from the classicality conditions. The data could very faithfully be represented in two-sector Fock space, thus providing support to our quantum-theoretic modeling. More, they allowed us to attain new fundamental results in concept research and to sustain and corroborate our explanatory hypothesis in Section \ref{explanation}. Hence, it is worth to briefly review our recent results starting from the conditions for classicality of conceptual data sets, i.e. representability of empirical membership weights in a 
 Kolmogorovian probability space.
 
Let $\mu(A), \mu(B), \mu(A'), \mu(B')$, $\mu(A\ {\rm and}\ B)$, $\mu(A\ {\rm and}\ B')$, $\mu(A'\ {\rm and}\ B)$, and $\mu(A'\ {\rm and}\ B')$ be the membership weights of an exemplar $x$ with respect to the concepts $A$, $B$, their negations `not $A$', `not $B$' and the conjunctions `$A$ and $B$', `$A$ and not $B$', `not $A$ and $B$' and `not $A$ and not $B$', respectively, and suppose that all these membership weights are  contained in the interval $[0,1]$ (which they will be in case they are experimentally determined as limits of relative frequencies of respective memberships). Then, they are `classical conjunction data' if and only if they satisfy the following conditions.
\begin{eqnarray} \label{condbis01}
&\mu(A)=\mu(A\ {\rm and}\ B)+\mu(A \ {\rm and}\ B') \\ \label{condbis02}
&\mu(B)=\mu(A\ {\rm and}\ B)+\mu(A' \ {\rm and}\ B) \\ \label{condbis03}
&\mu(A')=\mu(A'\ {\rm and}\ B')+\mu(A' \ {\rm and}\ B) \\ \label{condbis04}
&\mu(B')=\mu(A'\ {\rm and}\ B')+\mu(A \ {\rm and}\ B') \\ \label{condbis05}
&\mu(A\ {\rm and}\ B)+\mu(A\ {\rm and}\ B')+\mu(A'\ {\rm and}\ B)+\mu(A'\ {\rm and}\ B')=1
\end{eqnarray}
(see \cite{asv2014} for the proof).

A large amount of data collected in \cite{asv2014} violates very strongly and also very systematically (\ref{condbis01})--(\ref{condbis05}), hence these data cannot be generally reproduced in a classical Kolmogorovian probability framework. It can instead be shown that almost all these data can be represented by using our  quantum-theoretic modeling in two-sector Fock space, as above. For the sake of simplicity, let us work out separate representations for the two sectors.

Let us start from sector 1 of Fock space, which models genuine emergence. We represent the concepts $A$, $B$ and their negations `not $A$', `not $B$' by the mutually orthogonal unit vectors $|A\rangle$, $|B\rangle$ and $|A'\rangle$, $|B'\rangle$, respectively, in the individual Hilbert space ${\cal H}$. The corresponding membership weights for a given exemplar $x$ are then given by the quantum probabilistic Born rule
\begin{eqnarray}
\mu(A)=\langle A|M|A\rangle & \quad & \mu(B)=\langle B|M|B\rangle \\
\mu(A')=\langle A'|M|A'\rangle& \quad & \mu(B')=\langle B'|M|B'\rangle
\end{eqnarray}
in sector 1. The conjunctions `$A$ and $B$', `$A$ and not $B$', `not $A$ and $B$', and `not $A$ and not $B$' are represented by the superposition vectors $\frac{1}{\sqrt{2}}(|A\rangle+|B\rangle)$, $\frac{1}{\sqrt{2}}(|A\rangle+|B'\rangle)$, $\frac{1}{\sqrt{2}}(|A'\rangle+|B\rangle)$ and $\frac{1}{\sqrt{2}}(|A'\rangle+|B'\rangle)$, respectively, in ${\cal H}$, i.e. sector 1 of Fock space, which expresses the fact `$A$ and $B$', `$A$ and not $B$', `not $A$ and $B$', and `not $A$ and not $B$' are considered as newly emergent concepts in sector 1.

Let us come to sector 2 of Fock space, which models logical reasoning.  
Here we introduce a new element,
expressing an insight which we had not 
yet in our earlier application of Fock space \cite{s2014b,a2009a,ags2013,s2014a}, and which we explain in detail in \cite{asv2014}. In short it comes to `taking into account that possibly $A$ and $B$ are meaning-connected and hence their probability weights mutually 
dependent'. If this is the case, we cannot represent, e.g., the conjunction `$A$ and $B$' by the tensor product vector $|A\rangle \otimes |B\rangle$ of ${\cal H} \otimes {\cal H}$. This would indeed entail that the membership weight for the conjunction is $\mu(A \ {\rm and}  \ B)=\mu(A)\mu(B)$ in sector 2, that is, probabilistic independence between the membership estimations of $A$ and $B$. We instead, following this new insight, represent the conjunction `$A$ and $B$' by an arbitrary vector $|C\rangle\in {\cal H} \otimes {\cal H}$, in sector 2, which in general will be entangled if $A$ and $B$ are meaning dependent. If we represent the decision measurements of a subject estimating the membership of the exemplar $x$ with respect to the concepts $A$ and $B$ by the orthogonal projection operators $M\otimes \mathbbm{1}$ and $\mathbbm{1}\otimes M$, respectively, we have
\begin{equation}
\mu(A)=\langle C|M\otimes \mathbbm{1}|C\rangle \quad
\mu(B)=\langle C| \mathbbm{1}\otimes M|C\rangle
\end{equation}
in sector 2. We have now to formalize the fact that this sector 2 has to express logical relationships between the concepts. More explicitly, the decision measurements of a subject estimating the membership of the exemplar $x$ with respect to the negations `not $A$' and `not $B$' should be represented by the orthogonal projection operators $(\mathbbmss{1}-M)\otimes \mathbbm{1}$ and $\mathbbm{1}\otimes (\mathbbmss{1}-M)$, respectively, in sector 2, in such a way that
\begin{equation}
\mu(A')=1-\mu(A)=\langle C|(\mathbbm{1}-M)\otimes \mathbbm{1}|C\rangle \quad
\mu(B')=1-\mu(B)=\langle C| \mathbbm{1}\otimes (\mathbbm{1}-M|C\rangle)
\end{equation}
in this sector.

Interestingly enough, there is a striking connection between logic and classical probability when conjunction and negation of concepts are at stake. Namely, the logical probabilistic structure of sector 2 of Fock space sets the limits of classical probabilistic models, and  vice versa. In other words, if the experimentally collected membership weights $\mu(A)$, $\mu(B)$, $\mu(A')$, $\mu(B')$, $\mu(A\ {\rm and}\ B)$, $\mu(A \ {\rm and}\ B')$, $\mu(A'\ {\rm and}\ B)$ and $\mu(A'\ {\rm and}\ B')$ can be represented in sector 2 of Fock space for a given choice of the  state vector $|C\rangle$ and the decision measurement projection operator $M$, then the membership weights satisfy (\ref{condbis01})--(\ref{condbis05}), hence they are classical data.  Vice versa, if $\mu(A)$, $\mu(B)$, $\mu(A')$, $\mu(B')$, $\mu(A\ {\rm and}\ B)$, $\mu(A \ {\rm and}\ B')$, $\mu(A'\ {\rm and}\ B)$ and $\mu(A'\ {\rm and}\ B')$ satisfy (\ref{condbis01})--(\ref{condbis05}), hence they are classical data, then an entangled state vector $|C\rangle$ and a decision measurement projection operator $M$ can always be found such that $\mu(A)$, $\mu(B)$, $\mu(A')$, $\mu(B')$, $\mu(A\ {\rm and}\ B)$, $\mu(A \ {\rm and}\ B')$, $\mu(A'\ {\rm and}\ B)$ and $\mu(A'\ {\rm and}\ B')$ can be represented in sector 2 of Fock space (see \cite{asv2014} for the proof).

Let us finally come to the general representation in two-sector Fock space. We can now introduce the general form of the vector representing the state of the conjunction of the concepts $A, B$ and their respective negations.
\begin{eqnarray}
| \Psi_{AB} \rangle&=& m_{AB}e^{i \lambda_{AB}} |C\rangle + \frac{n_{AB}e^{i \nu_{AB}}}{\sqrt{2}} (|A\rangle+|B\rangle) \\
| \Psi_{AB'} \rangle&=& m_{AB'}e^{i \lambda_{AB'}} |C\rangle + \frac{n_{AB'}e^{i \nu_{AB'}}}{\sqrt{2}} (|A\rangle+|B'\rangle) \\
| \Psi_{A'B} \rangle&=& m_{A'B}e^{i \lambda_{A'B}} |C\rangle + \frac{n_{A'B}e^{i \nu_{A'B}}}{\sqrt{2}} (|A'\rangle+|B\rangle) \\
| \Psi_{A'B'} \rangle&=& m_{A'B'}e^{i \lambda_{A'B'}} |C\rangle + \frac{n_{A'B'}e^{i \nu_{A'B'}}}{\sqrt{2}} (|A'\rangle+|B'\rangle)
\end{eqnarray}
where $m^2_{XY}+n^2_{XY}=1$, $X=A,A',Y=B,B'$. 
The corresponding membership weights $\mu(A\ {\rm and}\ B)$, $\mu(A \ {\rm and}\ B')$, $\mu(A'\ {\rm and}\ B)$ and $\mu(A'\ {\rm and}\ B')$ can be written as in (\ref{AND}). We proved in \cite{asv2014} that they can be  expressed in the Fock space ${\mathbb C}^{8} \oplus ({\mathbb C}^{8} \otimes {\mathbb C}^{8})$ as
\begin{eqnarray}
\mu(A\ {\rm and}\ B)&=&m_{AB}^2 \alpha_{AB}+n_{AB}^2(\frac{\mu(A)+\mu(B)}{2}+\beta_{AB}\cos\phi_{AB})
\label{FockSpaceSolutionAB} \\
\mu(A \ {\rm and} \ B')&=&m_{AB'}^2 \alpha_{AB'}+n_{AB'}^2(\frac{\mu(A)+\mu(B')}{2}+\beta_{AB'}\cos\phi_{AB'}) \label{FockSpaceSolutionAB'} \\
\mu(A' \ {\rm and}\ B)&=&m_{A'B}^2 \alpha_{A'B}+n_{A'B}^2(\frac{\mu(A')+\mu(B)}{2}+\beta_{A'B}\cos\phi_{A'B})
\label{FockSpaceSolutionA'B} \\
\mu(A'\ {\rm and}\ B')&=&m_{A'B'}^2\alpha_{A'B'}+n_{A'B'}^2(\frac{\mu(A')+\mu(B')}{2}+\beta_{A'B'}\cos\phi_{A'B'}) \label{FockSpaceSolutionA'B'}
\end{eqnarray}
where $0\le \alpha_{XY}\le 1$, $-1\le \beta_{XY} \le 1$,  $X=A,A',Y=B,B'$.

Let us consider a relevant example, {\it Goldfish}, with respect to ({\it Pets}, {\it Farmyard Animals}) (big overextension in all experiments, but also double overextension with respect to {\it Not Pets And Farmyard Animals}). {\it Goldfish} scored $\mu(A)=0.93$ with respect to {\it Pets}, $\mu(B)=0.17$ with respect to {\it Farmyard Animals}, $\mu(A')=0.12$ with respect to {\it Not Pets}, $\mu(B')=0.81$ with respect to {\it Not Farmyard Animals}, $\mu(A \ {\rm and} \ B)=0.43$ with respect to {\it Pets And Farmyard Animals}, $\mu(A \ {\rm and} \ B')=0.91$ with respect to {\it Pets And Not Farmyard Animals}, $\mu(A' \ {\rm and} \ B)=0.18$ with respect to {\it Not Pets And Farmyard Animals}, and $\mu(A' \ {\rm and} \ B')=0.43$ with respect to {\it Not Pets And Not Farmyard Animals}. A complete modeling in the Fock space satisfying Eqs. (\ref{FockSpaceSolutionAB}), (\ref{FockSpaceSolutionAB'}), (\ref{FockSpaceSolutionA'B}) and (\ref{FockSpaceSolutionA'B'}) is characterized by coefficients:

(i) interference angles $\phi_{AB}=78.9^{\circ}$, $\phi_{AB'}=43.15^{\circ}$, $\phi_{A'B}=54.74^{\circ}$ and $\phi_{A'B'}=77.94^{\circ}$;

(ii) coefficents $\alpha_{AB}=0.12$, $\alpha_{AB'}=0.8$, $\alpha_{A'B}=0.05$ and $\alpha_{A'B'}=0.03$;

(iii) coefficients 
$\beta_{AB}=-0.24$, 
$\beta_{AB'}=0.10$, $\beta_{A'B}=0.12$ and $\beta_{A'B'}=0.30$;

(iv)  convex weights 
$m_{AB}=0.45$,
$n_{AB}=0.89$, 
$m_{AB'}=0.45$, 
$n_{AB'}=0.9$, 
$m_{A'B}=0.48$,  $n_{A'B}=0.88$, $m_{A'B'}=0.45$, and $n_{A'B'}=0.89$. 

Following our interpretation in the case of conjunction and disjunction, we can say that, whenever a subject is asked to estimate whether a given exemplar $x$ belongs to the concepts $A$, $B$, `$A \ {\rm and} \ {\rm not} \ B$', both quantum logical and quantum conceptual thought simultaneously act in the subject's thought. According to the former, the subject considers two copies of $x$ and estimates whether the first copy belongs to $A$ and the second copy of $x$ does not belong to $B$. According to the latter, the subject estimates whether the exemplar $x$ belongs to the newly emergent concept `$A \ {\rm and} \ {\rm not} \ B$'. Fock space naturally captures this two-layered structure.

\section{A unifying explanatory hypothesis\label{explanation}}
The Fock space modeling presented in the previous section suggested us to formulate a general hypothesis which justifies and explains a whole set of empirical results on cognitive psychology under a unifying theoretic scheme \cite{asvIQSA2}. According to our explanatory hypothesis, human reasoning is a specifically structured superposition of two processes, a `logical reasoning' and an `emergent reasoning'. Logical reasoning combines cognitive entities -- concepts, combinations of concepts, or propositions -- by applying the rules of logic, though 
generally in a probabilistic  way. Emergent reasoning enables instead formation of combined cognitive entities as 
newly emerging  entities -- in the case of concepts, new concepts, in the case of propositions, new propositions --
carrying new meaning, connected with the meaning of the component cognitive entities, but with a connection not defined by the algebra of logic. These two mechanisms act simultaneously in human thought during a reasoning process, the first one is guided by an algebra of `logic', the second one follows a mechanism of `emergence'. 

Human reasoning can be mathematically formalized in the two-sector Fock space presented in Section \ref{fockspace}. The states of conceptual entities are represented 
by unit vectors of this Fock space as we have seen in the specific case of concept combinations. More specifically,  `sector 1 of Fock space' models `conceptual emergence', hence the combination of two concepts is represented by a superposition vector of 
the vectors representing the component concepts in this Hilbert space, allowing `quantum interference' between conceptual entities
to play a role in the process of emergence. `Sector 2 of Fock space'  models a conceptual combination from the combining concepts by requiring 
the rules of logic  for the logical connective used for the combining, i.e. conjunction or disjunction, 
to be satisfied in a probabilistic setting. This quantum-theoretic modeling suggested us to call `quantum conceptual thought' the process occurring in sector 1 of Fock space, `quantum logical thought' the process occurring in sector 2. The relative importance of emergence or logic in a specific cognitive process is measured by the `degree of participation' of sectors 1 and 2. 

The abundance of evidence of deviations from classical logical reasoning in concrete human decisions (paradoxes, fallacies, effects, contradictions), together with our results in these two sections, led us to draw the conclusion that emergence constitutes the dominant dynamics of human reasoning, while logic is only a secondary structure. Therefore, we put forward the view that the aforementioned deviations from classicality are a consequence of the dominant dynamics and their nature is emergence, while classical logical reasoning is not a default to deviate from but, rather, a consequence of a secondary 
structure and its nature is logic.

There is further empirical evidence revealing that what primarily guides human subjects in concrete human decisions is emergent reasoning, but logical aspects are likewise present. 

A first element of evidence we identified by comparing the behavior of experimental data of different experiments. Consider, e.g., the exemplar {\it Olive} 
and its membership weights with respect to the concepts {\it Fruits}, {\it Vegetables} and their conjunction {\it Fruits And Vegetables}, measured by ourselves \cite{s2014b,asv2014}, and its membership weights with respect to the concepts {\it Fruits}, {\it Vegetables} and their disjunction {\it Fruits Or Vegetables}, measured by Hampton \cite{h1988b}. 
{\it Olive} scored $\mu(A)=0.56$ with respect to {\it Fruits}, $\mu(B)=0.63$ with respect to {\it Vegetables} and $\mu(A \ {\rm and} \ B)=0.65$ with respect to {\it Fruits 
And Vegetables}, that is, {\it Olive} was double overextended with respect to the conjunction.
But, {\it Olive} was also double overextended with respect to the disjunction, since it scored $\mu(A)=0.5$ with respect to {\it Fruits}, $\mu(B)=0.1$ with respect to {\it Vegetables} and $\mu(A \ {\rm 
or} \ B)=0.8$ with respect to {\it Fruits Or Vegetables}.
Our interpretation of these {\it Olive} case is the following. People see {\it Olive} as an exemplar which could be considered to be a fruit, but also could be considered to be a vegetable. Hence it could also, and even more so, be considered to be both, a fruit `and' a vegetable. This explains the double overextension of {\it Olive} with respect to the conjunction. This way of looking at {\it Olive} gives indeed the necessary weight to the conjunction to produce a double overextension.
Equally so, people see {\it Olive} as an exemplar which induces doubt about whether it is a vegetable or whether it is a fruit. Hence its could also, and even more so, be considered to be a 'fruit or a vegetable'. This explains the double overextension of {\it Olive} with respect to the disjunction. This way of looking at {\it Olive} gives indeed the necessary weight to the disjunction to produce a very big double overextension. Let us remark indeed that a double overextention with respect to the disjunction does not necessarily violates the classicality conditions, on the contrary, for a classical probability, the disjunction should be double overextended. For {\it Olive} the overextension is however so big that another one of the classicality conditions, namely the one linked to the Kolmogorovian factor is violated. For a classical probability model we have $\mu(A)+\mu(B)-\mu(A \ {\rm 
or} \ B)=\mu(A \ {\rm and}\ B)$, which means $0\le \mu(A)+\mu(B)-\mu(A \ {\rm 
or} \ B)$. However, for {\it Olive} we have $\mu(A)+\mu(B)-\mu(A \ {\rm 
or} \ B)=0.5+0.1-0.8=-0.2<0$, which shows that the double overextension for the disjunction in the case of {\it Olive} is of a non-classical nature.  
How come that {\it Olive} can give such weight to both conjunction and disjunction, although conjunction and disjunction are considered in classical probability to be distinctly different? It is because the meaning of {\it Olive} plays dominantly in sector 1, were quantum conceptual structures exist, and both connectives, `and' and `or' quite well resemble each other in this realm of conceptuality. 

The second empirical evidence became manifest when we calculated the deviations from (\ref{condbis01})--(\ref{condbis05}) across all exemplars in our experiment in \cite{s2014b,asv2014}, and we noticed that these deviations have approximately constant numerical values. Indeed, let us introduce the following quantities.
\begin{equation}
I_{ABA'B'}=1-\mu(A\ {\rm and}\ B)-\mu(A\ {\rm and}\ B')-\mu(A'\ {\rm and}\ B)-\mu(A'\ {\rm and}\ B') \label{normalization}
\end{equation}
\begin{eqnarray}
I_{A}&=&\mu(A)-\mu(A\ {\rm and}\ B)-\mu(A \ {\rm and}\ B') \label{negationA} \\
I_{B}&=&\mu(B)-\mu(A\ {\rm and}\ B)-\mu(A' \ {\rm and}\ B) \label{negationB}\\
I_{A'}&=&\mu(A')-\mu(A'\ {\rm and}\ B')-\mu(A' \ {\rm and}\ B) \label{negationA'}\\
I_{B'}&=&\mu(B')-\mu(A'\ {\rm and}\ B')-\mu(A \ {\rm and}\ B') \label{negationB'}
\end{eqnarray}
We were very excited ourselves to find that for every $X=A,A',Y=B,B'$, $I_{X}$, $I_{Y}$ and $I_{ABA'B'}$ are constant functions  across all exemplars, because this constitutes a very strong experimental evidence for the non-classical nature of what happens during concept combinations. More concretely, the last four equations give rise to values between $0$, which would be the classical value, and $-0.5$, but substantially closer to $-0.5$ than to $0$, and the fifth equation gives rise to a value between $0$, which again would be the classical value, and $-1$, but closer to $-1$ than to $0$. This is very strong evidence for the presence of non-classicality, indeed, if the classicality conditions are violated in such a strong and systematical way, the underlying structure cannot in any way be classical.
To test 
the rugularity of this violation we firstly performed a `linear regression analysis' of the data to check whether these quantities can be represented by a line of the form $y=mx+q$, with $m=0$. This was the case.  For $I_{A}$, we obtained $m=3.0 \cdot 10^{-3}$ with $R^{2}=0.94$; for $I_{B}$, we obtained $m=2.9 \cdot 10^{-3}$ with $R^{2}=0.93$; for $I_{A'}$, we obtained $m=2.6 \cdot 10^{-3}$ with $R^{2}=0.96$; for $I_{B'}$, we obtained $m=3.1 \cdot 10^{-3}$ with $R^{2}=0.98$; for $I_{ABA'B'}$, we obtained $m=4 \cdot 10^{-3}$ with $R^{2}=0.92$. Secondly, we computed the $95\%$-confidence interval for these parameters and obtained interval $(-0.51, -0.33)$ for $I_A$, interval $(-0.42, -0.28)$ for $I_{A'}$, interval $(-0.52, -0.34)$ for $I_B$, interval $(-0.40, -0.26)$ for $I_{B'}$, and interval $(-0.97, -0.64)$ for $I_{ABA'B'}$. This means that the measured parameters systematically fall within a narrow band centered at very similar values. 
Next to the very strong experimental evidence for the non-classical nature of the underlying structure, the finding of  this very stable pattern of violation constitutes also strong evidence for the validity of our Fock space model, and for the dominance of emergent reasoning with respect to logical reasoning when concepts are combined. Indeed, suppose for a moment that we substitute in the place of the experimental values in our equations to test classicality, the values that would be obtained theoretically in case we apply the first sector of Fock space equation of our Fock space model. Since interference in this equation can be negative as well as positive, and there is a priori no reason to suppose that there would be more of the one than the other, we can neglect the interference parts of the equation, since it is reasonable to suppose that they will cancel out when summing on all the terms of the equations of our classicality conditions. This means that we get the following, for every $X=A,A',Y=B,B'$, we have $\mu(X \ {\rm and } \ Y)=\frac{1}{2}({\mu(X)+\mu(Y)})$ (see   (\ref{FockSpaceSolutionAB})--(\ref{FockSpaceSolutionA'B'})). A simple calculation shows that, for every $X=A,A',Y=B,B'$, $I_{X}=I_{Y}=-0.5$ and $I_{ABA'B'}=-1$, in this case. These are exactly the values to which our experimental violations are close, which means that are Fock space model captures the underlying structures in a systematic and deep way. The experimental values are in between these values, and $0$. which is the classical value, which means that also logical reasoning is present, but the emergent reasoning is dominant.  

We think that these two results confirm, on one side, the general validity of our quantum-theoretic perspective in cognition and, on the other side, they constitute a very strong experimental support to the explanatory hypothesis presented in this section.

The next two sections complete our overview on the identification of quantum structures in concept combination, also shedding new light on the mysteries that surround quantum entanglement and indistinguishability at a microscopic level.

\section{Identification of entanglement\label{entanglement}}
The presence of entanglement  is typically revealed in quantum physics by a violation of Bell-type inequalities \cite{b1964,chsh1969}, indicating that the corresponding coincidence measurements exhibit correlations that cannot be modeled in a classical Kolmogorovian probability framework \cite{a1986,p1989}. 

We recently measured in a cognitive test statistical correlations in the conceptual combination {\it The Animal Acts}. We experimentally found that this combination violates Bell's inequalities \cite{ags2013,as2011} and elaborated a model that faithfully represents the collected data in complex Hilbert space \cite{as2014}. {\it The Animal Acts} unexpectedly revealed the presence of a `conceptual entanglement' which is only partly due to the component concepts, or `state entanglement', because it is also caused by `entangled measurements' and `entangled dynamical evolutions between measurements' \cite{as2014}. Our analysis shed new light on the mathematical and conceptual foundations of quantum entanglement, revealing that situations are possible where only states are entangled and measurements are products (`customary state entanglement'), but also situations where entanglement appears on the level of the measurements, in the form of the presence of both entangled measurements and entangled evolutions (`nonlocal box situation', `nonlocal non-marginal box situation'), due to the violation of the marginal distribution law, as in {\it The Animal Acts}.  More specifically, {\it The Animal Acts} is a paradigmatic example of a `nonlocal non-marginal box situation', that is, an experimental situation where (i) joint probabilities do not factorize, (ii) Bell's inequalities are violated, and (iii) the marginal distribution law does not hold. Whenever these conditions are simultaneously satisfied, a form of entanglement appears which is stronger than the `customarily identified quantum entanglement in the states of microscopic entities'. In these cases, it is not possible to work out a quantum-mechanical representation in
a fixed ${\mathbb C}^2\otimes{\mathbb C}^2$ space which satisfies empirical data and where only the initial state is entangled while the measurements are products. It follows that entanglement is a more complex property than usually thought. 
Shortly, if a single measurement is at play, one can distribute the entanglement between state and measurement but, if more measurements are considered, the marginal distribution law imposes limits on the ways to model the presence of the entanglement.

Let us now come to our coincidence measurements $e_{AB}$, $e_{AB'}$, $e_{A'B}$ and $e_{A'B'}$ for the conceptual combination {\it The Animal Acts}. In all measurements, we asked subjects to answer the question `is a good example of' the concept {\it The Animal Acts}. In measurement $e_{AB}$, participants choose among the four possibilities (1) {\it The Horse Growls}, (2) {\it The Bear Whinnies} -- and if one of these is chosen, the outcome is $+1$ -- and (3) {\it The Horse Whinnies}, (4) {\it The Bear Growls} -- and if one of these is chosen, the outcome is $-1$. In measurement $e_{AB'}$, they choose among (1) {\it The Horse Snorts}, (2) {\it The Bear Meows} -- and in case one of these is chosen, the outcome is $+1$ -- and (3) {\it The Horse Meows}, (4) {\it The Bear Snorts} -- and in case one of these is chosen,  the outcome is $-1$. In measurement $e_{A'B}$, they choose among (1) {\it The Tiger Growls}, (2) {\it The Cat Whinnies} -- and in case one of these is chosen, the outcome is $+1$ -- and (3) {\it The Tiger Whinnies}, (4) {\it The Cat Growls} -- and in case one of these is chosen, the outcome is $-1$. 
Finally, in measurement $e_{A'B'}$, participants choose among (1) {\it The Tiger Snorts}, (2) {\it The Cat Meows} -- and in case one of these is chosen,  the outcome is $+1$ -- and (3) {\it The Tiger Meows}, (4) {\it The Cat Snorts} -- and in case one of these is chosen,  the outcome is $-1$. We evaluate now the expectation values $E(A,B)$, $E(A, B')$, $E(A', B)$ and $E(A', B')$ associated with the measurements $e_{AB}$, $e_{AB'}$, $e_{A'B}$ and $e_{A'B'}$ respectively, and insert the values into the Clauser-Horne-Shimony-Holt (CHSH) version of Bell's inequality \cite{chsh1969} 
\begin{equation} \label{chsh}
-2 \le E(A',B')+E(A',B)+E(A,B')-E(A,B) \le 2
\end{equation}
We performed a test on 81 participants who were presented a questionnaire to be filled out in which they were asked to choose among the above alternatives in $e_{AB}$, $e_{AB'}$, $e_{A'B}$ and $e_{A'B'}$. 
Table 1 contains the results of our experiment \cite{as2011}.

If we denote by $P(H,G)$, $P(B,W)$, $P(H,W)$ and $P(B,G)$, the probability that {\it The Horse Growls}, {\it The Bear Whinnies},  
{\it The Horse Whinnies} and {\it The Bear Growls}, respectively, is chosen in $e_{AB}$, and so for in the other measurements, the expectation values are, in the large number limits, 
\begin{eqnarray}
&E(A,B)=&p(H,G)+p(B,W)-p(B,G)-p(H,W)=-0.7778  \nonumber \\
&E(A',B)=&p(T,G)+p(C,W)-p(C,G)-p(T,W)=0.6543 \nonumber \\
&E(A,B')=&p(H,S)+p(B,M)-p(B,S)-p(H,M)=0.3580 \nonumber \\
&E(A',B')=&p(T,S)+p(C,M)-p(C,S)-p(T,M)=  0.6296 \nonumber
\end{eqnarray}
Hence, (\ref{chsh}) gives 
\begin{equation}
E(A',B')+E(A',B)+E(A,B')-E(A,B)=2.4197
\end{equation}
which is significantly greater than 2. This implies that (i) it violates Bell's inequalities, and (ii) the violation is close the maximal possible violation in quantum theory, i.e. $2\cdot\sqrt{2} \approx 2.8284$.
\begin{table} \label{tab}
\centering
\begin{footnotesize}
\begin{tabular}{|c | c | c| c| }
\hline
 \emph{Horse Growls} & \emph{Horse Whinnies} & \emph{Bear Growls} & \emph{Bear Whinnies}\\
 $p(H,G)=0.049$ & $p(H,W)=0.630$ & $p(B,G)=0.259$ & $p(B,W)=0.062$  \\
\hline
  \emph{Horse Snorts} & \emph{Horse Meows} & \emph{Bear Snorts} & \emph{Bear Meows}\\
$p(H,S)=0.593$ & $p(H, M)=0.025$ & $p(B,S)=0.296$   & $p(B,M)=0.086$ \\
\hline
 \emph{Tiger Growls} & \emph{Tiger Whinnies} & \emph{Cat Growls} & \emph{Cat Whinnies}\\
  $p(T,G)=0.778$ & $p(T, W)=0.086$ & $p(C,G)=0.086$  &  $p(C,W)=0.049$ \\
\hline
 \emph{Tiger Snorts} & \emph{Tiger Meows} & \emph{Cat Snorts} & \emph{Cat Meows}\\
  $p(T,S)=0.148$ & $p(T, M)=0.086$ & $p(C,S)=0.099$ & $p(C,M)=0.667$\\
\hline
\end{tabular}
\caption{The data collected in coincidence measurements on entanglement in concepts \cite{as2011}.}
\end{footnotesize}
\end{table}

Let us now construct a quantum representation in complex Hilbert space for the collected data by starting from an operational description of the conceptual entity {\it The Animal Acts}. The entity {\it The Animal Acts} is abstractly described by an initial state $p$. Measurement $e_{AB}$ has four outcomes $\lambda_{HG}$, $\lambda_{HW}$, $\lambda_{BG}$ and $\lambda_{BW}$, and four final states $p_{HG}$, $p_{HW}$, $p_{BG}$ and $p_{BW}$. Measurement $AB'$ has four outcomes $\lambda_{HS}$, $\lambda_{HM}$, $\lambda_{BS}$ and $\lambda_{BM}$, and four final states $p_{HS}$, $p_{HM}$, $p_{BS}$ and $p_{BM}$. Measurement $A'B$ has four outcomes $\lambda_{TG}$, $\lambda_{CG}$, $\lambda_{TW}$ and $\lambda_{CW}$, and four final states $p_{TG}$, $p_{TW}$, $p_{CG}$ and $p_{CW}$. Measurement $A'B'$ has four outcomes $\lambda_{TS}$, $\lambda_{CS}$, $\lambda_{TM}$ and $\lambda_{CM}$, and four final states $p_{TS}$, $p_{TM}$, $p_{CS}$ and $p_{CM}$.
Then, we consider the Hilbert space ${\mathbb C}^4$ as the state space of {\it The Animal Acts} and represent the state $p$ by the unit vector $|p\rangle \in {\mathbb C}^4$. We assume that $\{|p_{HG}\rangle, |p_{HW}\rangle, |p_{BG}\rangle,$ $ |p_{BW}\rangle \}$, $\{|p_{HS}\rangle, |p_{HM}\rangle, |p_{BS}\rangle, |p_{BM}\rangle\}$, $\{|p_{TG}\rangle$, $ |p_{TW}\rangle$, $|p_{CG}\rangle$, $|p_{CW}\rangle\}$, $\{|p_{TS}\rangle,$ $|p_{TM}\rangle, |p_{CS}\rangle, |p_{CM}\rangle\}$ are orthonormal (ON) bases of ${\mathbb C}^4$. Therefore, $|\langle p_{HG}|\psi\rangle|^2=p(H,G)$, $|\langle p_{HW}|\psi\rangle|^2=p(H,W)$, $|\langle p_{BG}|\psi\rangle|^2=p(B,G)$, $|\langle p_{BW}|\psi\rangle|^2=p(B,W)$, in the measurement $e_{AB}$. We proceed analogously for the other probabilities. Hence, the self-adjoint operators 
\begin{eqnarray}
{\cal E}_{AB}&=&\sum_{i=H,B}\sum_{j=G,W}\lambda_{ij}|p_{ij}\rangle \langle p_{ij}| \\
{\cal E}_{AB'}&=&\sum_{i=H,B}\sum_{j=S,M}\lambda_{ij}|p_{ij}\rangle \langle p_{ij}|\\
{\cal E}_{A'B}&=&\sum_{i=T,C}\sum_{j=G,W}\lambda_{ij}|p_{ij}\rangle \langle p_{ij}|\\
{\cal E}_{A'B'}&=&\sum_{i=T,C}\sum_{j=S,M}\lambda_{ij}|p_{ij}\rangle \langle p_{ij}|
\end{eqnarray}
represent the measurements $e_{AB}$, $e_{AB'}$, $e_{A'B}$ and $e_{A'B'}$ in ${\mathbb C}^4$, respectively.

Let now the state $p$ of {\it The Animal Acts} be the entangled state represented by the unit vector $|p\rangle=|0.23e^{i13.93^\circ}, 0.62e^{i16.72^\circ},0.75e^{i9.69^\circ},0e^{i194.15^\circ}\rangle$ in the canonical basis of ${\mathbb C}^{4}$. This choice is not arbitrary, but deliberately `as close as possible to a situation of only product measurements', as we explained in \cite{as2014}. Moreover, we choose the outcomes $\lambda_{HG}=\lambda_{BW}=+1$, $\lambda_{HW}=\lambda_{BG}=-1$, and so on, as in our concrete experiment. We proved that 
\scriptsize
\begin{eqnarray}
{\cal E}_{AB}
&=&
\left( \begin{array}{cccc}
0.952 & -0.207-0.030i	&	0.224+0.007i & 0.003-0.006i \\				
-0.207+0.030i & -0.930	&	0.028-0.001i	&	-0.163+0.251i \\				
0.224-0.007i & 0.028+0.001i & -0.916 & -0.193+0.266i \\
0.003+0.006i & -0.163-0.251i & -0.193-0.266i & 0.895				
\end{array} \right)
\\
{\cal E}_{AB'}
&=&
\left( \begin{array}{cccc}
-0.001	&				0.587+0.397i &	0.555+0.434i &	0.035+0.0259i	\\			
0.587-0.397i & -0.489 &	0.497+0.0341i &	-0.106-0.005i \\	
0.555-0.434i & 0.497-0.0341i & -0.503	&	0.045-0.001i \\				
0.035-0.0259i &	-0.106+0.005i &	0.045+0.001i & 0.992	\end{array} \right)
\\
{\cal E}_{A'B}
&=&
\left( \begin{array}{cccc}
-0.587 &	0.568+0.353i	&	0.274+0.365i	&	0.002+0.004i \\																							
0.568-0.353i & 0.090	 &				0.681+0.263i & -0,110-0.007i \\				
0.274-0.365i &		0.681-0.263i & -0.484	&	0.150-0.050i \\				
0,002-0.004i & -0,110+0.007i & 0.150+0.050i &	0.981				
\end{array} \right)
\end{eqnarray}
\begin{eqnarray}
{\cal E}_{A'B'} 
&=&
\left( \begin{array}{cccc}
0.854	&				0.385+0.243i & -0.035-0.164i &	-0.115-0.146i \\				
0.385-0.243i & -0.700	&	0.483+0.132i & -0.086+0.212i \\				
-0.035+0.164i &	0.483-0.132i & 0.542 &	0.093+0.647i \\				
-0.115+0.146i &	-0.086-0.212i &	0.093-0.647i &	-0.697	
\end{array} \right)
\end{eqnarray}
\normalsize
in this case \cite{as2014}. 

This completes the quantum-theoretic modeling in ${\mathbb C}^{4}$ for our cognitive test. One can then resort to the definitions of entangled states and entangled measurements and to the canonical isomorphisms, ${\mathbb C}^{4}\cong{\mathbb C}^{2}\otimes{\mathbb C}^{2}$ and $L({\mathbb C}^{4})\cong L({\mathbb C}^{2})\otimes L({\mathbb C}^{2})$ ($L({\cal H})$ denotes the vector space of linear operators on the Hilbert space $\cal H$), and one can prove that all measurements $e_{AB}$, $e_{AB'}$, $e_{A'B}$ and $e_{A'B'}$ are entangled with this choice of the entangled state $p$  \cite{as2014}. Moreover, the marginal distribution law is violated by all measurements, e.g., $p(H,G)+p(H,W) \ne p(H,S)+p(H,M)$. Since we are below Tsirelson's bound \cite{tsirelson80}, this modeling is an example of a `nonlocal non-marginal box modeling 1', following the classification we proposed in \cite{asQI2013}.

To conclude the section we remind that we have used the term `entanglement' by explicitly referring to the structure within the theory of quantum physics that a modeling of experimental data takes, if (i) these data are represented, following carefully the rules of standard quantum theory, in a complex Hilbert space, and hence states, measurements, and evolutions, are presented respectively by vectors (or density operators), self-adjoint operators, and unitary operators in this Hilbert space; (ii) a situation of coincidence joint measurement on a compound entity is considered, and the subentities are identified following the tensor product rule of `compound entity description in quantum theory' (iii) within this tensor product description of the compound entity entanglement is identified, as `not being product', whether it is for states (non-product vectors), measurements (non-product self-adjoint operators), or evolutions (non-product unitary transformations).

\section{The quantum nature of conceptual identity\label{identity}}
One of the most mysterious and less understood aspects of quantum entities is the way they behave with respect to `identity', and more specifically their statistical behaviour due to indistinguishability. Indeed, the statistical behaviour of quantum entities is very different from the statistical behaviour of classical objects, which are instead, in principle, 
not identical, hence distinguishable, whenever there is more than one. The latter is governed by the Maxwell-Boltzmann (MB) distribution, while the former is described by the Bose-Einstein (BE) distribution for quantum particles with integer spin, and by the Fermi-Dirac (FD) distribution for quantum particles with semi-integer spin (we omit considering fractional statistical particles here, for the sake of brevity) \cite{French2006,Dieks2008}. 

What about concepts? Consider, e.g., the linguistic expression ``eleven animals''. This expression, when both ``eleven'' and ``animals'' are looked upon with respect to their conceptual structure, represents the combination of concepts {\it Eleven} and {\it Animals} into {\it Eleven Animals}, which is again a concept. 
Each of the {\it Eleven Animals} is then completely identical on this conceptual level, and hence indistinguishable. The same linguistic expression can  however also elicit the thought about eleven objects, present in space and time, each of them being an instantiation of {\it Animal}, and thus distinguishable from each other.  We recently inquired into experiments on such combinations of concepts, surprisingly finding that BE statistics appears at an empirical level for specific types of concepts, hence finding strong evidence for the hypothesis that indeed there is a profound connection between the behavior of concepts with respect to identity and indistinguishability and the behaviour of quantum entities with respect to these notions \cite{asvIQSA1}.
What is interesting in this respect is that we can intuitively understand the behavior of concepts with respect to identity and indistinguishability, which means that it might well be that an understanding of the behavior of quantum entities with respect to identity and indistinguishability should be searches for by making use of this analogy. In this sense, that identical concepts can be modeled exactly as identical quantum entities, i.e. by using quantum theory, is not only a strong achievement for quantum cognition, but it might also incorporate a new way to reflect about this mysterious behaviour of identical quantum entities.

Let us discuss these aspects both at a theoretic and an empirical level, as follows.

Let us firstly consider the SCoP structure in Section \ref{contextuality} and two states of {\it Animal}, namely {\it Cat} and {\it Dog}, hence the situation where {\it Eleven Animals} can be either {\it Cats} or {\it Dogs}. Then, the conceptual meaning of {\it Eleven Animals}, which can be {\it Cats} or {\it Dogs}, gives rise in a unique way to twelve possible states. Let us denote them by $p_{11,0}$, $p_{10,1}$, \ldots, $p_{1,10}$ and $p_{0,11}$, and they stand respectively for {\it Eleven Cats} (and no dogs), {\it Ten Cats And One Dog}, \ldots, {\it One Cat And Ten Dogs} and {\it Eleven Dogs} (and no cats). 
We investigated the `probabilities of change of the ground state $\hat p$ of the combined concept {\it Eleven Animals} into one of the twelve states $p_{11,0}$, $p_{10,1}$, \ldots, $p_{1,10}$ and $p_{0,11}$' in a cognitive experiment on human subjects. The subjects were presented the twelve states and asked to choose their preferred one. The relative frequency arising from their answers was interpreted as the probability of change of the ground state $\hat p$, to the chosen state, i.e. one of the set $\{p_{11,0}$, $p_{10,1}$, \ldots, $p_{1,10}$, $p_{0,11}\}$. The context $e$ involved in this experiment is mainly determined by the  
`combination procedure of the concepts {\it Eleven} and {\it Animals}'  
and the `meaning contained in the new combination' for participants in the experiment \cite{ag2005a}. Hence, our psychological experiment tested whether participants follow the `conceptual meaning' of {\it Eleven Animals} treating {\it Dogs} ({\it Cats}) as identical, or participants follow the `instantiations into objects meaning' of {\it Eleven Animals} treating {\it Dogs} ({\it Cats}) as distinguishable.

We mathematically represent the conceptual entity {\it Eleven Animals} by the SCoP model $(\Sigma,{\cal M},
\mu)$, where  $\Sigma=\{\hat p, p_{11,0}, p_{10,1}, \ldots, p_{1,10}, p_{0,11}\}$, ${\cal M}= \{ e \}$, and our transition probabilities are $\{\mu(q,e,\hat p)\ \vert q\in \{p_{11,0}, p_{10,1}, \ldots, p_{1,10}, p_{0,11}\}\}$. We recognise in the structure of $\mu(q,e,\hat p)$ the situation is analogous to the one in which one has $N=11$ particles that can be distributed in $M=2$ possible states. It is 
thus possible, by looking at the relative frequencies obtained in the experiment, to find out whether a classical MB statistics or a quantum-type, i.e. BE statistics, applies to this situation. In case that MB would apply, it would mean that things happen as if there are underlying the twelve states hidden possibilities, namely $T(n,C;11-n,D)=11!/n!(11-n)!$ in number, for the specific state of $n$ {\it Cats} and $11-n$ {\it Dogs}, $n=0, \ldots,11$. Of course, this ``is'' true in case the cats and dogs are real cats and dogs, hence are `objects existing in space and time', which is why for objects in the classical world indeed MB statistics applies. Let us calculate the probabilities involved then. 
For the sake of simplicity, we assume that two probability values $P_{Cat}$ and $P_{Dog}$ exist such that $P_{Cat}+P_{Dog}=1$, and that the events of 
making actual such an underlying state for {\it Cat} and {\it Dog} are independent. Hence the probability 
for $n$ exemplars of {\it Cat} and $11-n$ exemplars of {\it Dog} is
then 
\begin{equation}
\mu_{MB}^{P_{Cat},P_{Dog}}(p_{n,11-n},e,\hat p)=T(n,C;11-n,D) P_{Cat}^nP_{Dog}^{11-n}=\frac{11!}{n!(11-n)!} P_{Cat}^nP_{Dog}^{11-n}
\end{equation}
 Note that, under  the assumption of MB statistics, $\mu_{MB}^{P_{Cat},P_{Dog}}(p,e,\hat p)$ becomes the binomial probability distribution. For example, if $P_{Dog}=P_{Cat}=0.5$, the number of possible arrangements for the state {\it Eleven Cats And Zero Dogs} and for the state {\it Zero Cats And Eleven Dogs} is 1, hence the corresponding probability for these configurations is $\mu_{MB}^{P_{Cat},P_{Dog}}(p_{0,11},e,\hat p)=\mu_{MB}^{P_{Cat},P_{Dog}}(p_{11,0},e,\hat p)=0.0005$. Analogously, the number of possible arrangements for the state {\it Ten Cats And One Dog} and for the state {\it One Cat And Ten Dogs} is 11, hence the corresponding probability for these configurations is $\mu_{MB}^{P_{Cat},P_{Dog}}(p_{10,1},e,\hat p)=\mu_{MB}^{P_{Cat},P_{Dog}}(p_{1,10},e,\hat p)=0.0054$, and so on. When $P_{Cat}$ and  $P_{Dog}$ are  equal, MB distribution entails a maximum value for such a probability. In this example, this corresponds to the situation of {\it Six Cats And Five Dogs} and {\it Five Cats And Six Dogs} with $\mu_{MB}^{P_{Cat},P_{Dog}}(p_{6,5},e,\hat p)=\mu_{MB}^{P_{Cat},P_{Dog}}(p_{6,5},e,\hat p)=0.2256$. 

Let us now make the calculation for BE statistics, where we keep making the exercise of only reasoning on the level of concepts, and not on the level of instantiations. This means that the twelve different states do not admit underlying hidden states, because the existence of such states would mean that we reason on more concrete forms in the direction of instantiations.  As above, we suppose that {\it Cat} and {\it Dog} have an independent elicitation probability  $P_{Cat}$ and $P_{Dog}$ such that $P_{Cat}+P_{Dog}=1$. Hence, the probability that there are $n$ exemplars of {\it  Cat} and $(11-n)$ exemplars of {\it Dog} is 
\begin{equation}
\mu_{BE}^{P_{Cat},P_{Dog}}(p_{n,11-n},e,\hat p)= \frac{(nP_{cat}+(11-n)P_{dog})}{(\frac{12\times 11}{2})}
\end{equation}
Note that as $P_{Cat}=1-P_{Dog}$, then $\mu_{BE}^{P_{Cat},P_{Dog}}(p_{n,11-n},e,\hat p)$ is a linear function. Moreover, when 
$P_{Cat}=P_{Dog}=0.5$, we have that $\mu_{BE}(p_{n,11-n},e,\hat p)=1/12$ for all values of $n$, thus recovering BE distribution \cite{asvIQSA1}.

Starting from the above theoretic analysis, if one performs experiments on a collection of concepts like {\it Eleven Animals} to estimate the probability of elicitation for each state, then one can establish whether a distribution of MB-type $\mu_{MB}^{P_{Cat},P_{Dog}}(p_{n,11-n},e,\hat p)$, or of BE-type $\mu_{BE}^{P_{Cat},P_{Dog}}(p_{n,11-n},e,\hat p)$, or a different one, holds. However, in case there are strong deviations from a MB statistics, while a quasi-linear distribution is obtained, then this would indicate that, in context $e$, where only {\it Cat} and {\it Dog} are allowed to be states of the concept {\it Animal}, the statistical distribution of the collection of concepts {\it Eleven Animals} is of a BE-type and that concepts present a quantum-type indistinguishability.

We performed a cognitive experiment with 88 participants. We considered a list of concepts  $A^i$ of different (physical and non-physical) nature, $i=1, \ldots,14$, and two possible exemplars (states) $p_{1}^i$ and $p_{2}^i$ for each concept. Next we requested participants to choose one exemplar of a combination $N^{i} A^i$ of concepts, where 
$N^{i}$ is a natural number. The exemplars of these combinations of concepts 
$A^i$ are the states $p^i_{k,N^{i}-k}$ describing the conceptual combination   `$k$ exemplars in state $p^i_1$  and $(N^i-k)$ exemplars in state $p^i_2$', where $k$ is an integer such that 
$k=0, \ldots, N^i$. For example, the first collection of concepts we considered is  $N^1 A^1$ corresponding to the compound conceptual entity {\it Eleven Animals}, with $p^i_{1}$ and $p^i_{2}$ describing the exemplars  {\it Cat} and {\it Dog} of the individual concept {\it Animal}, respectively, and  $N^1=11$. The exemplars (states) we considered are thus $p^1_{11,0}$, $p^1_{10,1}$, \ldots, $p^{1}_{1,10}$, and $p^{1}_{0,11}$, describing the combination  {\it Eleven Cats And Zero Dogs}, {\it Ten Cats And One Dog}, \ldots, {\it One Cat And Ten Dogs}, and  {\it  Zero Cats And Eleven Dogs}. The other collections of concepts we considered in our cognitive experiment are reported in Table 2.
\begin{table}[H]\label{categories-psych}
\begin{center} 
\begin{tabular}{|c|c|c|c|c|} \hline
$i$ & $N^i$ & $A^i$ & $p_{1}^i$ & $p_{2}^i$  \\ 
\hline 
\hline
1& 11 & {\it Animals} & {\it Cat} & {\it Dog} \\ \hline
2& 9 & {\it Humans} & {\it Man} & {\it Woman}\\ \hline
3& 8 & {\it Expressions of Emotion} & {\it Laugh} & {\it Cry} \\ \hline
4& 7& {\it Expressions of Affection} & {\it Kiss} & {\it Hug}\\ \hline
5& 11& {\it Moods} & {\it Happy} & {\it Sad} \\ \hline
6& 8 &{\it Parts of Face} & {\it Nose} & {\it Chin} \\ \hline 
7& 9 & {\it Movements} & {\it Step} & {\it Run} \\ \hline
8& 11 & {\it Animals} & {\it Whale} & {\it Condor}\\ \hline
9& 9 & {\it Humans} & {\it Child} & {\it Elder} \\ \hline
10& 8 & {\it Expressions of Emotion} & {\it Sigh} & {\it Moan} \\ \hline
11& 7& {\it Expressions of Affection} & {\it Caress} & {\it Present}\\ \hline
12& 11& {\it Moods} & {\it Thoughtful} & {\it Bored}\\ \hline
13& 8 &{\it Parts of Face} &{\it Eye} & {\it Cheek} \\ \hline
14& 9 &{\it Movements} & {\it Jump} & {\it Crawl} \\ \hline
\end{tabular}
\caption{List of concepts and their respective states for the psychological concept on identity and indistinguishability.}
\end{center} 
\end{table}

We computed the parameters $P^{MB}_{p_{1}^i}$ and $P^{BE}_{p_{1}^i}$ that minimize the the R-squared value of the fit using the  distributions $\mu_{MB}^{P_{p_{1}^i},P_{p_{2}^i}}$ and $\mu_{BE}^{P_{p_{1}^i},P_{p_{2}^i}}$ for each $i=1, \ldots, 14$. Hence, we fitted the distributions obtained in the psychological experiments using MB and BE statistics 
(note that only one 
parameter is needed as $P^{MB}_{p_{2}^i}=1-P^{MB}_{p_{1}^i}$ and $P^{BE}_{p_{2}^i}=1-P^{BE}_{p_{1}^i}$).
Next, we used the `Bayesian Information Criterion (BIC)'  \cite{KASSBIC} to estimate which model provides the best fit and contrast this criterion with the R-squared value.
 Table
3 summarizes the statistical analysis. The first column of the table identifies the concept in question (see Table 2),
the second and third columns show $P^{MB}_{p_{1}^i}$ and the $R^2$ value of the MB statistical fit, the fourth and fifth columns show $P^{BE}_{p_{1}^i}$ and the $R^2$ value of the BE statistical fit. The sixth column shows the $\Delta_{\textrm{BIC}}$ criterion to discern between the $\mu_{MB}^{P_{p_{1}^i}P_{p_{2}^i}}$ and $\mu_{BE}^{P_{p_{1}^i}P_{p_{2}^i}}$, and the seventh column identifies the distribution which best fits the data for concept  $A^i$, $i=1, \ldots,14$.
\begin{table}[H]
\begin{center} 
\begin{tabular}{|c|c|c|c|c|c|c|c|} \hline
$i$&$P^{MB}_{p_{1}^i}$&$R_{MB}^2$&$P^{BE}_{p_{1}^i}$&$R_{BE}^2$&$\Delta_{\textrm{BIC}}$ &Best Model  \\ \hline \hline
1& 0.55 &-0.05 &0.16 &0.78 &19.31  & BE strong \\ \hline
2& 0.57 &{\bf 0.78} &0.42 &0.44 &-9.54 &MB strong \\ \hline
3& 0.82 &0.29 &{\bf 0.96} &0.79 & 10.81&BE strong \\ \hline
4& 0.71 &0.81 &0.53 &0.77 &-1.69 & MB weak\\ \hline
5& 0.25 &{\bf 0.79} &0.39 &0.93 & 14.27&BE strong\\ \hline
6& 0.62 &0.59 &0.61 &0.57 &-0.37 &MB weak\\ \hline
7& 0.72 &0.41 &0.64 &{\bf 0.83} &12.66 &BE strong\\ \hline
8& 0.63 &0.58 &0.47 &0.73 &5.53 & BE positive\\ \hline
9& 0.45 &{\bf 0.87} &0.26 &0.67 &-9.69 &MB strong\\ \hline
10&0.59 &0.50 &0.63 &0.77 &7.17 &BE positive\\ \hline
11&0.86 &0.46 &1.00 &{\bf 0.87} &11.4 &BE strong \\ \hline
12&0.21 &0.77 &0.00 &0.87 &6.68 &BE positive\\ \hline
13&0.62 &0.54 &0.71 &0.67 &2.97 &BE weak \\ \hline
14&0.81 &0.20 &0.91 &{\bf 0.90} &20.68 &BE strong\\ \hline
\end{tabular}
\label{stat-results}
\caption{Results of statistical fit for the psychological experiment. Each column refers to the 14 collections of concepts introduced in Table 2.}
\end{center} 
\end{table}
Note that, according to the BIC criteria, negative $\Delta_{\textrm{BIC}}$ values imply that the category is best fitted by a MB distribution, whereas positive $\Delta_{\textrm{BIC}}$ values on row $i$ imply the concept 
$A^i$ is best fitted with a BE distribution. Moreover, when $|\Delta_{\textrm{BIC}}|<2$ there is no clear difference between the models, when $2<|\Delta_{\textrm{BIC}}|<6$ we can establish a positive but not strong difference towards the model with smallest value, whereas when $6<|\Delta_{\textrm{BIC}}|$ we are in presence of strong evidence that one of the models provides better fit than the other model~\cite{KASSBIC}. We see that categories $2$ and $9$ show a strong $\Delta_{\textrm{BIC}}$ value towards MB-type of statistics, and that categories $1,3,5,7,11,12$ and $14$ show a strong $\Delta_{\textrm{BIC}}$ value towards BE-type of statistics.
Complementary to the BIC criterion, the $R^2$ fit indicator helps to see whether or not the indications of $\Delta_{\textrm{BIC}}$ can be confirmed with a good fit of the data. Interestingly, the concepts we have identified with strong indication towards one type of statistics have $R^2$ values larger than $0.78$ (such $R^2$ values are marked in bold text), which indicates a fairly good approximation for the data. Moreover, note that in all the cases with strong tendency towards one type of statistics, the $R^2$ of the other type of statistics shows is poor. This confirms the fact that we can discern between the two types of statistics depending on the concept in question.

The interpretation of our results is thus clear. Conceptual combinations exists, like {\it Nine Humans}, whose distribution follows MB statistics. However, also conceptual combinations, like {\it Eleven Animals}, {\it Eight Expressions of Emotion} or {\it Eleven Moods}, whose distribution follows BE statistics exist. The conclusion is that the nature of identity in these concept combinations is of a quantum-type and 
in these combinations the human mind treats the two states we consider as identical and indistinguishable.  Also the hypothesis that `the more easy the human mind 
imagines spontaneously instantiations, e.g., {\it Nine Humans}, the more MB, and the less easy such instantiations are activated in imagination, e.g., {\it Eight Expressions of Emotion}, the more BE statistics appears' is confirmed by our experiment.

We have an intuitive explanation for this empirical difference. Whenever the human mind `imagines' two different combinations of {\it Eleven Animals}, say 2 cats and 9 dogs, and 5 cats and 6 dogs, the human mind does not really take into account that this situation can be about real cats and dogs, in which case there are many more ways to put 5 cats and 6 dogs into a cage, than to put 2 cats and 9 dogs. This is the reason why BE, not MB, appears in this case. Suppose instead that the human mind considers two different combinations of {\it Nine Humans}, say 2 elders and 7 children, and 4 elders and 5 children. Then  the human mind is likely to be influenced by real known families with 9 sons and, in the real world, there are much more situations of families with 4 elders and 5 children, than 2 elders and 7 children. 

This pattern was confirmed by a second experiment we performed on the World-Wide Web about the nature of conceptual indistinguishability \cite{asvIQSA1}.

\section{Concluding remarks and perspectives\label{conclusions}}
In the previous sections we have provided an overview of our quantum-theoretic perspective on concepts and their combinations. We have expounded the reasons that led us to develop this perspective, namely our former research on operational and axiomatic approaches to quantum physics, the origins of quantum probability and various experimental results in cognition pointing to a deviation of human reasoning from the structures of classical (fuzzy set) and classical probability theory. We have proved that these deviations can be described in terms of genuine quantum effects, such as contextuality, emergence, interference and superposition. We have identified further quantum aspects in the mechanisms of conceptual combination, namely entanglement and quantum-type indistinguishability. And, we have proposed an explanation that allows the unification of these different empirical results under a common underlying principle on the structure of human thought. 

Our quantum-theoretic perspective fits the global research domain that applies the mathematical formalisms of quantum theory in cognitive science and has been called `quantum cognition' (we quote here some known books on this flourishing domain \cite{k2010,bb2012,hk2013}). Further, we believe that our findings in cognition may also have, as a feedback, a deep impact on the foundations of microscopic quantum physics. Indeed, let us consider entanglement. We identified in concepts an entanglement situation where Bell's inequalities are violated within Tsirelson's bound \cite{tsirelson80}, the marginal distribution law is violated and there is `no signaling', which implies that entangled measurements, in addition to entangled states, are needed to model this experimental situation. And this completely occurs within a Hilbert space quantum framework, at variance with widespread beliefs. This theoretic scheme with entangled measurements could explain some `anomalies', i.e. deviations from the marginal distribution law, that were recently observed in the typical Bell-type nonlocality tests with entangled photons \cite{ak1,ak2}. Then, let us consider the quantum nature of conceptual indistinguishability. In our perspective, it is due to 
the human mind  being able to consider specific conceptual entitites 
without the need to also imagine instantiations as objects existing in space and time of these conceptual entities. 
Hence, it could well be that quantum indistinguishability at a microscopic level 
is provoked by the fact that quantum entitites 
are not localized as objects in space and time, 
and that non-locality would mean non-spatiality, a view that has been put forward by one of us in earlier work for different reasons \cite{aerts1990}. 
These insights could for example have implications on quantum statistics and the so-called `spin-statistics theorem' \cite{French2006,Dieks2008}.

We conclude this article with an epistemological consideration. We think that our quantum-theoretic perspective in concept theory constitutes a step towards the construction of a general theory for 
the modeling of conceptual entities. In this sense, we distinguish it from what is typically considered as 
an ad hoc cognitive model. To understand what we mean by this distinction let us consider an example taken from everyday life. As an example of a theory, we could introduce the theory of `how to make good clothes'. A tailor needs to learn how to make good clothes for different types of people, men, women, children, old people, etc. Each cloth is a model on itself. Then, one can also consider intermediate situations where one has models of series of clothes. A specific body will not fit in any clothes: you need to adjust the parameters (length, size, etc.) to reach the desired fit. We think that 
a theory 
should be able to reproduce different experimental results by suitably adjusting the involved parameters, exactly as a 
theory of clothing. This is 
different from a set of ad hoc models, even if the set can cope with a wide range of experimental data. There is a tendency in psychology to be critical for a theory that can cope with all possible situations it applies to. One then often believes that the theory contains too many parameters, and that it is only by allowing all these parameters to attain different values that all the data can be modeled. In case we have to do with an ad hoc model, i.e, a model specially made for the circumstance of the situation it models, this suspicion is grounded. Adding parameters to such an ad hoc model, or stretching the already contained parameters to other values, does not give rise to what we call a theory.
A theory needs to be well defined, its rules, the allowed procedures, its theoretical, mathematical, and internal logical structure, `independent' of the structure of the models describing specific situations that can be coped with by the theory. Hence also the theory needs to contain a well defined description of `how to produce models for specific situations'. Think again of the theory of clothing. If a taylor knows the theory of clothing, obviously he or she can make a cloth for every human body, because the theory of clothing, although its structure is defined independently of a specific clothe, contains a prescription of how to apply it to any possible specific cloth. Other subtle aspects are involved with the differences between ad hoc models and models finding their origin in a theory \cite{aertsrohrlich1998}. Here we mainly wanted to bring up the issue, because we think it does lead to misunderstandings not to pay attention to the difference between an ad hoc model, and a model which is derived from a theory. Intuitive thoughts about the nature of a model differ often depending on whether the model is inspired by a psychology approach, it will then rather automatically been looked upon as an ad hoc model, and that all data can be modelled is the suspicious, or whether it is inspired by a physics approach, where it will rather be looked upon as resulting from a theory, and that all data can be modeled by it is the a positive aspect, validating the theory.

What is the status of the Fock space model for concept combinations? Hilbert space, hence also Fock space when appropriate, for the description of quantum entities provides models that definitely come from a theory, namely quantum theory, and hence are not ad hoc models. Is quantum theory also a theory for concepts and their combinations, and hence, if so, can we consider our models, e.g. the Fock space model, as models coming from this theory? Or is quantum cognition rather still a discipline where ad hoc models are built, making use, also in a rather ad hoc way, of mathematics arising from quantum theory? An answer to this question can not yet been given definitely, but some hypothesis can be formulated with plausibility in respect to it.   We believe that, notwithstanding their deep analogies, concept entities are less crystalized and symmetric structures than quantum entities. As a matter of fact some data in \cite{h1988a,h1988b} and \cite{s2014b,asv2014} cannot be modeled in Fock space, and further experimental findings could in the future confirm such impossibility. Notwithstanding this, we believe that emergence is the actual driver also for these data that cannot be  modeled in Fock space. But, this type of emergence cannot be represented in a linear Hilbert (Fock) space, and more general structures are needed. The search for more general mathematical structures capturing conceptual emergence will, by the way, constitute an important aspect of our future investigation in concept theory.
On the other hand, we do believe that we have arrived in the realm of building models that come from a theory, and are not ad hoc. Indeed, although we believe that this theory will turn out to be a generalisation of the actual quantum theory, its basic principles -- except linearity most probably -- will be present in the generalised quantum theory too. We have recently worked out an analysis where the view of the status of actual quantum cognition, as describing a quantum-like domain of reality, less crystallised than the micro-world, but containing deep analogies in its foundations, is put forward \cite{as2014b}.

\bigskip
\noindent
{\bf Acknowledgements.} We are greatly indebted to the scientists who have collaborated with us in these years on the fashinating research illustrated in the present article, namely, Sven Aerts, Jan Broekaert, Bart D'Hooghe, Marek Czachor, Liane Gabora, Massimiliano Sassoli de Bianchi, Sonja Smets, Jocelyn Tapia and Tomas Veloz. The results expounded here would not have been possible without our collaborations with them and their contributions.


\begin{thebibliography}{99}
\bibitem{r1973} Rosch, E.: Natural categories. Cogn. Psychol. \textbf{4}, 328--350 (1973)

\bibitem{z1989} Zadeh, L.A.: Knowledge representation in fuzzy logic. IEEE Transactions on Knowledge and Data Engineering 1, 89-100 (1989)

\bibitem{os1981} Osherson, D., Smith, E.: On the adequacy of prototype theory as a theory of concepts. Cognition \textbf{9}, 35--58 (1981)

\bibitem{h1988a} Hampton, J.A.: Overextension of conjunctive concepts: Evidence for a unitary model for concept typicality and class inclusion. J. Exp. Psychol. Learn. Mem. Cogn. \textbf{14}, 12--32 (1988a)

\bibitem{h1988b} Hampton, J.A.: Disjunction of natural concepts. Mem. Cogn. \textbf{16}, 579--591 (1988b)

\bibitem{s2014b} Sozzo, S.: Conjunction and negation of natural concepts: A quantum-theoretic modeling. J. Math. Psychol. \emph{http://dx.doi.org/10.1016/j.jmp.2015.01.005} (2015, in print) 

\bibitem{asv2014} Aerts, D., S. Sozzo, S., Veloz, T.:  Negation of natural concepts and the foundations of human reasoning (in preparation) 

\bibitem{h1997} Hampton, J.A.: Conceptual combination: Conjunction and negation of natural concepts. Mem. Cogn. \textbf{25}, 888--909 (1997)




\bibitem{ap2011} Alxatib, S., Pelletier, J.: On the psychology of truth gaps. In Nouwen, R., van Rooij, R., Sauerland, U., \& Schmitz, H.-C. (Eds.), {\it Vagueness in Communication} (pp. 13--36). Berlin, Heidelberg: Springer-Verlag (2011)

\bibitem{a2009a} Aerts, D.: Quantum structure in cognition. J. Math. Psychol. \textbf{53}, 314--348 (2009a)

\bibitem{ags2013} Aerts, D., Gabora, L., Sozzo, S.: Concepts and their dynamics: A quantum--theoretic modeling of human thought. Top. Cogn. Sci. \textbf{5}, 737--772 (2013)

\bibitem{s2014a} Sozzo, S.:  A quantum probability explanation in Fock space for borderline contradictions. J. Math. Psychol. \textbf{58}, 1--12 (2014)

\bibitem{a1986} Aerts, D.: A possible explanation for the probabilities of quantum mechanics. J. Math. Phys. \textbf{27}, 202--210 (1986)

\bibitem{aa1995} Aerts, D., Aerts, S.: Applications of quantum statistics in psychological studies of decision processes. Found. Sci. \textbf{1}, 85--97 (1995)

\bibitem{ag2005a} Aerts, D., Gabora, L.: A theory of concepts and their combinations I: The structure of the sets of contexts and properties. Kybernetes \textbf{34}, 167--191 (2005a)

\bibitem{ag2005b} Aerts, D., Gabora, L.: A theory of concepts and their combinations II: A Hilbert space representation. Kybernetes \textbf{34}, 192--221 (2005b)

\bibitem{a2009b} Aerts, D.: Quantum particles as conceptual entities: A possible explanatory framework for quantum theory. Foundations of Science \textbf{14}, 361--411 (2009b)

\bibitem{abgs2013} Aerts, D., Broekaert, J., Gabora, L., Sozzo, S.: Quantum structure and human thought. Behav. Bra. Sci. \textbf{36}, 274--276 (2013)

\bibitem{asvIQSA2} Aerts, D., S. Sozzo, S., Veloz, T.: Quantum structure in cognition and the foundations of human reasoning.  Archive reference and link: \emph{arXiv:} \emph{1412.8704} (2014)

\bibitem{as2011} Aerts, D., Sozzo, S.: Quantum structure in cognition. Why and how concepts are entangled. Lecture Notes in Computer Science vol. \textbf{7052}, 116--127 (2011)

\bibitem{as2014} Aerts, D., Sozzo, S. Quantum entanglement in conceptual combinations. Int. J. Theor. Phys. \textbf{53}, 3587--3603 (2014)

\bibitem{asvIQSA1} Aerts, D., S. Sozzo, S., Veloz, T.: The Quantum Nature of Identity in Human Thought: Bose-Einstein Statistics for Conceptual Indistinguishability. Archive reference and link: \emph{arXiv:} \emph{1410.6854} (2014)

\bibitem{abs1999} Aerts, D., Broekaert, J., Smets, S.:  A quantum structure description of the liar paradox. Int. J. Theor. Phys. \textbf{38}, 3231--3239 (1999)

\bibitem{aabg2000} Aerts, D., Aerts, S., Broekaert, J., Gabora, L.: The violation of Bell inequalities in the macroworld. Found. Phys. \textbf{30}, 1387--1414 (2000)


\bibitem{k1933} Kolmogorov, A.N.: Foundations of Probability. New York: Chelsea Publishing Company (1950)

\bibitem{p1989} Pitowsky, I.: Quantum Probability, Quantum Logic. Lecture Notes in Physics vol. \textbf{321}.  Berlin: Springer, Berlin (1989)

\bibitem{d1958} Dirac, P.A.M.: Quantum mechanics, 4th ed. Oxford University Press, Oxford (1958)

\bibitem{thisvolume} Haven, E., Khrennikov, A.Y.: A brief introduction to quantum formalism. In this volume (2015) 

\bibitem{b1964} Bell, J.S.: On the Einstein-Podolsky-Rosen paradox. Physics. \textbf{1}, 195--200 (1964)

\bibitem{chsh1969} Clauser, J.F., Horne, M.A., Shimony, A., Holt, R.A.: Proposed experiment to test local hidden-variable theories. Phys. Rev. Lett. \textbf{23}, 880--884 (1969)

\bibitem{tsirelson80} Tsirelson, B.S.: Quantum generalizations of Bell's inequality. Lett. Math. Phys. \textbf{4}, 93 (1980).

\bibitem{asQI2013} Aerts, D., Sozzo, S.: Entanglement Zoo I \& II. Lecture Notes in Computer Science vol. \textbf{8369}, 84--96 \& 97--109 (2014)
\bibitem{aerts1990} Aerts, D.: An attempt to imagine parts of the reality of the micro-world. In J. Mizerski, A. Posiewnik, J. Pykacz and M. Zukowski (Eds.), Problems in Quantum Physics (pp. 3-25). Singapore: World Scientific (1990).





\bibitem{French2006} French, S., Krause, D.: Identity in Physics: A Historical, Philosophical, and Formal Analysis. Oxford University Press, Oxford (2006)

\bibitem{Dieks2008} Dieks, D., Versteegh, M.A.M.: Identical quantum particles and weak discernibility. Found. Phys. {\bf 38}, 923-934 (2008)

\bibitem{aertsrohrlich1998} Aerts, D. and Rohrlich, F.: Reduction. Foundations of Science, 3, pp. 27-35, (1998).


\bibitem{as2014b} Aerts, D., \& Sozzo, S. (2014). What is quantum? Identifying its micro-physical and structural appearence. Accepted for publication in {\it Quantum Interaction. Lecture Notes in Computer Science}.


\bibitem{KASSBIC} Kass, R.E., Raftery, A.E.: Bayes factors. J. Am. Stat. Ass. \textbf{90}(430), 773--795 (1995)




\bibitem{k2010} Khrennikov, A. Y.: Ubiquitous Quantum Structure. Springer, Berlin (2010)

\bibitem{bb2012} Busemeyer, J.R., Bruza, P.D.: Quantum Models of Cognition and Decision. Cambridge University Press, Cambridge (2012)

\bibitem{hk2013} Haven, E., Khrennikov, A.Y.: Quantum Social Science. Cambridge University Press, Cambridge (2013)

\bibitem{ak1} Adenier, G., Khrennikov, A. Yu.: Anomalies in EPR-Bell Experiments. In Quantum Theory: Reconsideration of Foundations 3, Adenier, G., Khrennikov, A. Yu.,  Nieuwenhuizen, T. (eds.), pp. 283--293 (AIP, New York, 2006)

\bibitem{ak2} Adenier, G., Khrennikov, A. Yu.: Is the fair sampling assumption supported by EPR experiments? J. Phys A \textbf{40}, 131--141 (2007)  




\end{thebibliography}
\end{document}